%% file: main.tex
\documentclass[10pt,twocolumn,letterpaper]{article}

\usepackage{cvpr}              % To produce the CAMERA-READY version
\usepackage{graphicx}
\usepackage{amsmath}
\usepackage[table]{xcolor}
\usepackage{amssymb}
\usepackage{booktabs}
\usepackage{threeparttable}
\usepackage{pifont}
\usepackage{balance}
\usepackage{times}
\usepackage{epsfig}
\usepackage{algorithm}
\usepackage{algpseudocode}
\usepackage{anyfontsize}
\usepackage{url}
\usepackage{color}
\usepackage{caption}
\usepackage{mathrsfs}
\usepackage{float}
\usepackage{multirow}
\usepackage{pifont}
\usepackage{diagbox}
\usepackage[accsupp]{axessibility}
\definecolor{lightgray}{rgb}{0.95,0.95,0.95}
\usepackage[pagebackref,breaklinks,colorlinks]{hyperref}

\usepackage[capitalize]{cleveref}
\crefname{section}{Sec.}{Secs.}
\Crefname{section}{Section}{Sections}
\Crefname{table}{Table}{Tables}
\crefname{table}{Tab.}{Tabs.}

\def\cvprPaperID{6757} % *** Enter the CVPR Paper ID here
\def\confName{CVPR}
\def\confYear{2022}

\begin{document}

\title{Towards More Accurate Fake Detection on Images \\ Generated from Advanced Generative and Neural Rendering Models}

\newcommand{\email}[2]{\href{mailto:#1}{\textcolor{black}{#2}}}
\author{Chengdong~Dong\,$^{1,2}$\quad Vijayakumar~Bhagavatula\,$^{1}$ \quad Zhenyu~Zhou\,$^2$ \quad Ajay~Kumar\,$^{2}$\\
$^1$\,Department of Electrical and Computer Engineering, Carnegie Mellon University \\
$^2$\,Department of Data Science and Artificial Intelligence, The Hong Kong Polytechnic University \\
{\tt\small\{\email{chengdong.dong@connect.polyu.hk}{chengdong.dong},\email{zhenyucs.zhou@connect.polyu.hk}{zhenyucs.zhou}\}@connect.polyu.hk,\email{kumar@ece.cmu.edu}{kumar@ece.cmu.edu},\email{ajay.kumar@polyu.edu.hk}{ajay.kumar@polyu.edu.hk}}
}

\maketitle

\input{Sections/abstract}

\input{Sections/introduction}

\input{Sections/related_works}

\input{Sections/method}

\input{Sections/theory}

\input{Sections/datasets}

\input{Sections/experiments}

\input{Sections/conclusion}

\clearpage
%%%%%%%%% REFERENCES
\balance

{\small
\bibliographystyle{ieee_fullname}
\bibliography{references}
}

\clearpage
\renewcommand\appendixname{Supplementary}
\renewcommand\thefigure{\Alph{figure}}
\renewcommand\thetable{\Roman{table}}
\setcounter{figure}{0}
\setcounter{table}{0}
\appendix
\noindent 

\input{Sections/appendix_contents}

\end{document}

%% file: Sections/abstract.tex
%%%%%%%%% ABSTRACT
\begin{abstract}
The remarkable progress in neural-network-driven visual data generation, especially with neural rendering techniques like Neural Radiance Fields and 3D Gaussian splatting, offers a powerful alternative to GANs and diffusion models. These methods can produce high-fidelity images and lifelike avatars, highlighting the need for robust detection methods. In response, an unsupervised training technique is proposed that enables the model to extract comprehensive features from the Fourier spectrum's magnitude, thereby overcoming the challenges of reconstructing the spectrum due to its centrosymmetric properties. By leveraging the spectral domain and dynamically combining it with spatial domain information, we create a robust multimodal detector that demonstrates superior generalization capabilities in identifying challenging synthetic images generated by the latest image synthesis techniques. To address the absence of a 3D neural rendering-based fake image database, we develop a comprehensive database that includes images generated by diverse neural rendering techniques, providing a robust foundation for evaluating and advancing detection methods.
\end{abstract}

%% file: Sections/introduction.tex
%%%%%%%%% BODY TEXT
\vspace{-2mm}
\section{Introduction}
Images synthesized by generative models, such as Generative Adversarial Networks (GANs) \cite{progan,kwon2021diagonal,karras2021alias} and Diffusion Models (DM) \cite{guided,ldm,glide,li2024distrifusion}, have raised significant ethical, privacy and security related concerns in our society. A critical aspect overlooked by prior research is the potential for a large volume of generated images, if undetected, to contaminate the image pools available online, which are often used to train large-scale models. As noted in \cite{shumailov2024ai}, recursively generated data can lead to model collapse due to such contamination. To address such concerns emerging from the generative models, numerous neural synthetic image detection models \cite{wang2020cnn,ojha2023towards,moe_lora,durall2020watch,zhang2019detecting,liu2024forgery} have been developed. 

However, the advancement of neural rendering technologies such as Neural Radiance Fields (NeRF) \cite{ingp,tensorf,tancik2023nerfstudio,seathru,pynerf,sketchfacenerf,gsgen} and 3D Gaussian Splatting (3DGS) \cite{3dgs,c3dgs,splattingavatar} offers a novel approach to generating highly realistic imagery such as scenes and digital humans/avatars, by the acquisition of two-dimensional projections from lifelike three-dimensional spatial representations. There even exist methodologies \cite{instruct_n2n,instruct_gs2gs} capable of directly editing the content within 3D representations. Unlike generative models, neural rendering technologies produce more realistic synthetic images by reconstructing scenes from actual images, thereby avoiding logical and semantic inconsistencies. This process allows for subtle 3D modifications that, when projected to 2D, are nearly imperceptible. Notably, current fake image detection systems have not addressed whether neural-rendered images can also be identified as \textit{non-real}. This limitation prompts the question of whether current fake detectors possess sufficient generalization capabilities to detect neural-rendered images as fake. Specifically, when such detectors are trained exclusively on synthetic images produced by generative models and then tested on neural-rendered images, their efficacy remains uncertain.

To enhance generalization ability, \cite{ojha2023towards} uses linear probing to fine-tune the linear layer after a pre-trained large vision model. \cite{liu2024forgery,moe_lora} introduce side blocks to the main branch of fixed parameters inspired by LoRA \cite{hu2022lora} to improve representing ability. \cite{liu2024forgery} further enhances cross-domain performance by incorporating a frequency module (FAA) within the LoRA-like side blocks, while maintaining a spatial-based main branch. However, the representing power of these frequency modules is limited by the constrained scale of parameters.

In this work, we introduce Fourier Frequency-based image Transformer (FFiT), an architecture leveraging large vision models to extract spectral domain information for fake detection. The FFiT backbone is pre-trained using magnitudes of spectra from real images in an unsupervised way. Traditional Masked Autoencoders \cite{mae} struggle with spectral magnitudes due to their centrosymmetric properties (to be detailed later), leading to superficial reconstructions by copying patterns from unmasked symmetric patches and failing to recover the patch of magnitude from its neighboring unmasked patches. To address this, a novel loss function is introduced, enabling unsupervised learning of deep representations from spectral magnitudes. The multimodal architecture combines the pre-trained FFiT backbone with a spatial-based large vision model. As a binary classifier, it is fine-tuned on datasets containing both real and synthetic images. Both branches exhibit strong independent performance, attributable to their design and fine-tuning strategies. Specifically, our multimodal detector achieves average precision (AP) of 92.81\% and AUROC of 91.19\% across 11 types of 3D scene generators, training on real and GAN-generated images. It outperforms the state-of-the-art (SOTA) \cite{liu2024forgery}, proving that neural-rendered fake images can also be accurately detected.

Reference \cite{ojha2023towards} finds that detectors based on smaller neural networks, such as ResNet, struggle to identify challenging fake samples (e.g., DM-generated images) when trained on easily distinguishable fakes (e.g., GAN-generated images). In contrast, detectors using larger networks like ViT generalize well to difficult fakes even when trained on easier ones. Reference \cite{ricker2022towards} notes that training with difficult fake samples improves generalization for detectors, even those based on smaller networks. However, neither study offers a systematic explanation for these observations. Thus, an index is introduced to quantitatively analyze the hyperplane differentiating clustered features while mitigating the impact of sparse outliers, unveiling the relationship between cross-domain performance and the representing ability of deep models for fake detection.

To advance the research on the detection of neural-rendered images, we have collected a large-scale database comprising 296,504 images generated using NeRF and 3DGS, supplemented by 330,073 images produced via 3D scene editing techniques leveraging NeRF and 3DGS. Additionally, the advent of sophisticated video generation methods, such as Sora \cite{sora}, which demonstrate exceptional capability in displaying consistent 3D scenes, poses new challenges for the detection of synthetic visual content. Therefore, we also acquired 60,531 frames generated by Sora.

In summary, the main contributions are:

\noindent$\bullet$ Development of FFiT, the \textit{first} architecture using a large vision model to extract spectral domain imprints for accurately detecting fake images. This success is achieved by introducing a novel loss function to address the challenges posed by the spectrum's centrosymmetric property during unsupervised training.
 
\noindent $\bullet$ The dynamic consolidation of FFiT features with those from the spatial branch can achieve SOTA performance. Each network branch exhibits robust performance capabilities, which are attributed to its well-designed architecture and refined fine-tuning strategy.

\noindent$\bullet$ In contrast to existing databases that focus exclusively on fake images from generative models like GAN and DM, our database is the \textit{first} in its inclusion of images derived from neural rendering-based synthesized and edited 3D scenes, as well as realistic scenes generated by Sora.

\noindent$\bullet$ To analyze the generalization ability of detectors, a real-fake separation index is developed to quantitatively uncover the correlation between the cross-domain performance and representing the generalization ability of deep models for fake detection.

%% file: Sections/related_works.tex
\vspace{-2mm}
\section{Related Works}
\noindent\textbf{Realistic 3D Scene Generation}:
NeRF methods are developed to implicitly learn the 3D representation of specific scenes, which can be reconstructed from a series of input images \cite{ingp,tensorf,seathru,pynerf} or from a prompt such as a textual description \cite{dreamfusion} or a single-view image \cite{gao2024contex}. A potential malicious application involves integrating NeRF with editing techniques to alter the representation of 3D scenes using textual instructions \cite{instruct_n2n,jung2024geometry}, images \cite{pix2nerf,sketchfacenerf}, or both \cite{he2024customize}. Additionally, NeRF-based methods are used for generating realistic digital humans (avatars), such as speech-to-video talking heads \cite{adnerf,dfrf,geneface,genefacepp} and body synthesis \cite{ma2024humannerf,gao2024contex}.

The 3DGS methods \cite{3dgs,c3dgs} enable the learning of explicit 3D representations of scenes, which can be seamlessly integrated into existing rendering pipelines. Conditional editing techniques, such as Instruct-GS2GS \cite{instruct_gs2gs}, have been developed to modify 3DGS scenes. Avatar synthesis methods \cite{gaussianhead,splattingavatar} based on 3DGS have also been proposed. Moreover, 3DGS can be combined with diffusion models to generate 3D scenes from scratch, as exemplified by GSGEN \cite{gsgen} and DreamGaussian \cite{dreamgaussian}. Sora \cite{sora}, which is representative among a series of text-to-video generation methods \cite{sora,hong2022cogvideo,bar2024lumiere}, demonstrates an impressive ability to generate videos with accurate 3D relationships. This capability suggests that it possesses the capacity to represent the 3D world effectively. In this work, the focus is on detecting fake images synthesized by the aforementioned 3D scene generation methods.

\noindent\textbf{Fake Image Detection}:
To detect images generated by generative models, traditional detectors based on the spatial domain \cite{wang2020cnn,patchforensics,CoOccurrence,marra2019gans,zhao2021learning,tan2023learning} and those based on the spectral domain \cite{zhang2019detecting,dzanic2020fourier,durall2020watch,tan2024frequency} are trained on real and fake images to identify the latent fingerprints of GANs and Deepfakes. Methods that require learning \cite{DIRE,cazenavette2024fakeinversion,zhou2024stealthdiffusion}, and a learning-free method \cite{ricker2024aeroblade}, exploit inherent properties of DM architecture for detecting DM-generated fakes. Several approaches \cite{doloriel2024frequency,li2024masksim,NPR,ojha2023towards,moe_lora,liu2024forgery,cozzolino2024raising} are effective in identifying both GAN- and DM-generated images. Methods \cite{doloriel2024frequency,li2024masksim} enhance detection through an attention mechanism in the spectral domain. \cite{cozzolino2024raising} utilizes captions from real images to generate fakes and then trains an SVM using deep features from both real and generated fakes. NPR \cite{NPR} develops an operator to reveal neighboring pixel relationships within the spatial domain for improved detection. Ojha \etal \cite{ojha2023towards} introduce a large vision model with a fixed backbone to improve generalization, while \cite{liu2024forgery,moe_lora} fine-tune the backbone of large vision models like LoRA \cite{hu2022lora} to enhance representation while maintaining generalization. FatFormer \cite{liu2024forgery} further integrates information from frequency and language domains to boost cross-domain performance.

However, the aforementioned fake image detectors fail to address the challenge that the latent patterns of synthetic images generated by newer methods, such as NeRF and 3DGS, may significantly differ from those produced by traditional generative models due to the domain gap between these generation processes. Consequently, we introduces a novel architecture to address these emerging techniques and their associated detection challenges.

%% file: Sections/method.tex
\vspace{-2mm}
\section{Proposed Method}
In this section, we present FFiT, a framework that utilizes large vision models to extract spectral domain information for detecting fake images. The FFiT backbone is initially pre-trained in an unsupervised manner using magnitude spectra from real images. This pre-trained backbone is then combined with a spatial-based large vision model to create a multimodal architecture, which is fine-tuned on a dataset containing both real and synthetic images. The following subsections detail the architecture and training strategy of the proposed method.

\vspace{-2mm}
\subsection{Motivation and Design of FFiT}
\begin{figure}[ht]
  \centering
  \hspace{-6.7mm}
  \begin{subfigure}{0.244\linewidth}
  \centering
  \includegraphics[width=0.72in]{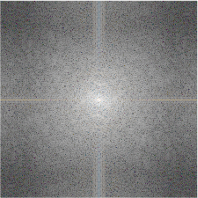}
    \caption{}
    \label{fig:mae_origin1}
  \end{subfigure}
  \begin{subfigure}{0.244\linewidth}
  \centering
\includegraphics[width=0.72in]{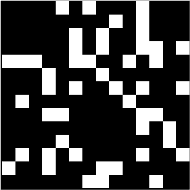}
    \caption{}
    \label{fig:mae_mask1}
  \end{subfigure}
    \begin{subfigure}{0.244\linewidth}
  \centering
\includegraphics[width=0.72in]{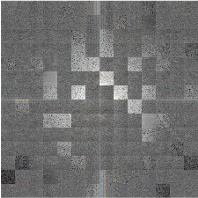}
    \caption{}
    \label{fig:mae_recon1}
  \end{subfigure}
  \\
    \begin{subfigure}{0.81\linewidth}
  \centering
\includegraphics[width=3in]{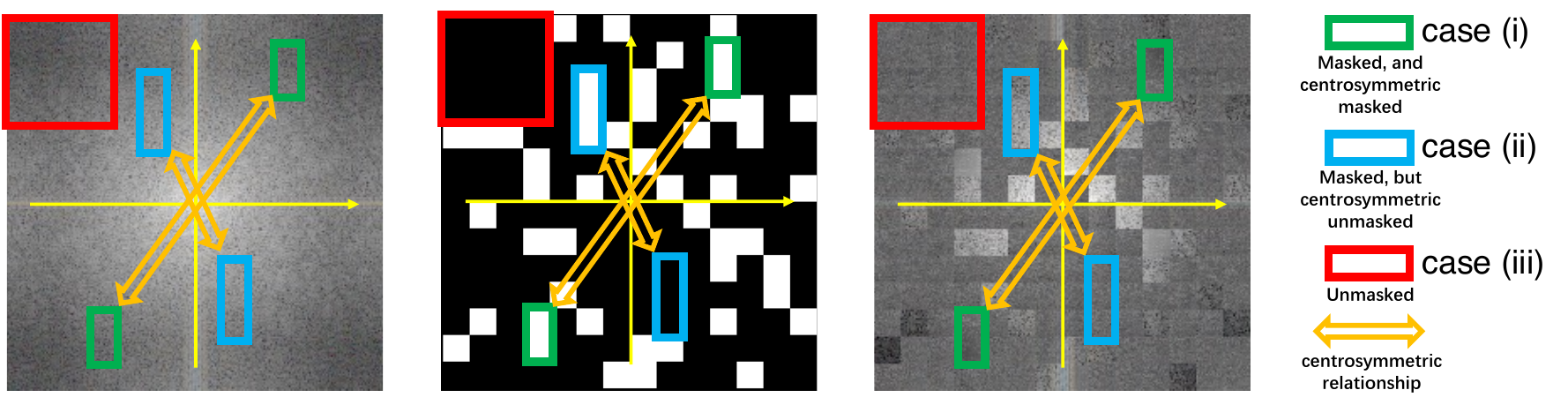}
    \caption{}
    \label{fig:mae_symmetric}
  \end{subfigure}
  \vspace{-2mm}
\caption{Failure in spectral information extraction with the original MAE pre-training. (a) Input spectrum magnitude. (b) Patch embedding mask for inference. (c) Poor-quality reconstruction from (a) and (b). (d) Explanation of reconstructed patches in (c).}
  \label{fig:mae1}
  \vspace{-4mm}
\end{figure}

MAE \cite{mae} is a classical method to train large neural models in an unsupervised way. However, the centrosymmetric characteristic of the spectrum, wherein the amplitudes at positive frequencies are equivalent to those at the corresponding negative frequencies, thereby exhibiting symmetry about the zero frequency, can introduce adverse effects to the training if we use the same way as MAE to train the Transformer on the spectral domain. In \cref{fig:mae_origin1}, a sample of the original magnitude of the spectrum is presented. \Cref{fig:mae_mask1} displays the mask utilized for patch masking during the inference phase, with the model that is trained using the original MAE-based training strategy. In this mask, white blocks indicate the patches that are to be masked during the patch embedding process, whereas black blocks represent the regions that should remain unmasked. \cref{fig:mae_recon1} illustrates the reconstructed magnitude of the spectrum, based on the input from \cref{fig:mae_origin1} and the mask shown in \cref{fig:mae_mask1}, demonstrating a poor quality of reconstruction. In \cref{fig:mae_symmetric}, three representative types of regions of \cref{fig:mae_recon1} are highlighted, and the following observations can be made: 

\noindent\textbf{case (\romannumeral 1):} When both a masked patch and its centrosymmetric counterpart are masked, the pre-trained model is unable to accurately reconstruct either. This indicates a limitation in the model's ability to infer information from the neighboring patches. 

\noindent\textbf{case (\romannumeral 2):} In the case of masked patches for which the corresponding centrosymmetric patches remain unmasked, the pre-trained model demonstrates a capability to reconstruct these patches with high accuracy. This suggests that the model effectively captures and utilizes the centrosymmetric property of the spectrum during training. 

\noindent\textbf{case (\romannumeral 3):} For the unmasked regions, it is evident that the pre-trained model fails to reconstruct them accurately. This finding is contrary to the expected behavior in the spatial domain, where an MAE-trained model typically succeeds in reconstructing unmasked areas.

\vspace{-2mm}
\subsubsection{Balancing the Weights of Various Masking Types}
In the original MAE training process, the block-wise reconstruction loss \(\mathcal{L}_{B(i,j)}\), which represents the reconstruction error for the \(i^{\text{th}}\) row and \(j^{\text{th}}\) column block (\(0 \leq i, j \leq N-1\)) between the original input magnitude of spectrum \(X\) and the reconstructed \(X'\), is calculated as follows:

\begin{small}
\vspace{-4mm}
\begin{equation}
\mathcal{L}_{B(i,j)} \!\!=\!\!\! \sum_{m=0}^{W-1} \! \sum_{n=0}^{W-1} \!\! || X \! (Wi \!+\! m, \! Wj \!+\! n \!) \!-\! X'\!(Wi \!+\! m, \! Wj \!+\! n\!)||^2
\vspace{-2mm}
\end{equation}
\end{small}
where $X$ is divided into \(N \times N\) patches (\(N\) is an even number) in a Transformer-based architecture. Given that \(X\) is of size \(224 \times 224\) pixels and each patch is of size \(W \times W\) pixels with \(W = 16\), we have \(N = 224 / W = 14\). During the training process, masks are applied to these patches, compelling the model to reconstruct the patterns within the masked regions, thereby facilitating unsupervised learning.

The total loss function in the original MAE training is computed by summing the reconstruction losses over the masked blocks, i.e., those \(B(i,j)\) that are masked. This approach has two key limitations that contribute to the failure to reconstruct the magnitude of the spectrum: 1. Ignorance of unmasked blocks in the loss function: the model is not penalized for any inaccuracies in the unmasked regions, which can lead to a lack of refinement in the overall reconstruction quality. 2. Overlooking centrosymmetric information: a masked block may have an unmasked centrosymmetric counterpart from which information can be easily copied. These limitations highlight the need for a more sophisticated loss function or training strategy that takes into account the unmasked regions and leverages the inherent symmetries within the spectral data to improve the reconstruction performance.

To address the limitations of the original MAE training process, we adopt a modified loss function that incorporates the focal loss mechanism. This approach aims to balance the influence of different masking cases, considering the special properties of the spectral magnitude. The loss function, denoted as \( \mathop{\mathcal{L}}_{r \neq 0}(X, X' | r)\), is defined as follows:

\begin{small}
\vspace{-4mm}
\begin{equation}
\hspace{-4mm}
        \mathop{\mathcal{L}}_{r \neq 0}(X, X' | r) =  -\frac{1}{N^2} \sum_{i=0}^{N-1} \sum_{j=0}^{N-1} \alpha_t (1 - \mathcal{L}_{B(i,j)})^\gamma \log \mathcal{L}_{B(i,j)}, \\
    \label{eq:focal_loss}
    \vspace{-2mm}
\end{equation}
\end{small}
for $B(i, j)$ $\in$ masking case $t$ $(t\in\{1,2,3\})$, $\alpha_t$ is used to balance its weight according to the occurring frequency of the specific case.

\begin{figure}[H]
  \centering
  \begin{subfigure}{0.24\linewidth}
  \centering
  \includegraphics[width=0.72in]{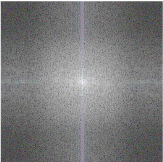}
    \caption{input}
    \label{fig:mae_origin2}
  \end{subfigure}
  \begin{subfigure}{0.24\linewidth}
  \centering
\includegraphics[width=0.72in]{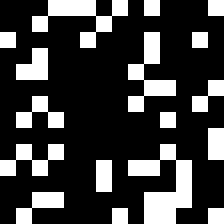}
    \caption{mask}
    \label{fig:mae_mask2}
  \end{subfigure}
    \begin{subfigure}{0.24\linewidth}
  \centering
\includegraphics[width=0.72in]{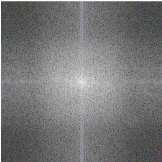}
    \caption{reconstructed}
    \label{fig:mae_recon2}
  \end{subfigure}
    \begin{subfigure}{0.24\linewidth}
  \centering
\includegraphics[width=0.72in]{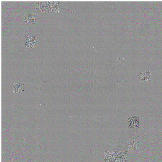}
    \caption{difference}
    \label{fig:mae_diff2}
  \end{subfigure}
  \vspace{-2mm}
  \caption{During training, we adopt the loss function in \cref{eq:focal_loss} and fix $r$ to 0.3. For inference, the mask ratio of (b) is set to 0.25.}
  \label{fig:mae2}
  \vspace{-2mm}
\end{figure}

The focusing parameter \( \gamma \) is designed to impose a greater penalty on hard examples. By increasing the value of \( \gamma \), the model places more emphasis on less frequent samples, thus improving their representation. The balancing factor \( \alpha_t \) is used to adjust the contribution of each class based on its effective number of samples. It is calculated as follows:
\begin{small}
\vspace{-2mm}
    \begin{equation}
        \alpha_t = \frac{1/P_t}{1/P_1 + 1/P_2 + 1/P_3}, \quad t=1,2,3
        \vspace{-2mm}
    \end{equation}
\end{small}
where \(P_1\), \(P_2\), and \(P_3\) refer to the expected probability for masking cases 1, 2, and 3.

The probability of a patch being masked is assumed to be \(r\). The expected number of pairs of masked blocks for the three different cases, which are denoted as \(E_1\), \(E_2\), and \(E_3\), can be computed as:

\begin{small}
\vspace{-4mm}
    \begin{equation}
        E_1 = \frac{N^2}{2} \!\times r^2, E_2 = \frac{N^2}{2} \!\times 2r \!\times (1 - r), E_3 = \frac{N^2}{2} \!\times (1 - r)^2
        \vspace{-2mm}
    \end{equation}
\end{small}

Thus the probabilities \(P_1\), \(P_2\), and \(P_3\) for three cases are:

\begin{small}
\vspace{-4mm}
    \begin{equation}
        P_1 = r^2, \quad P_2 = 2r \times (1 - r), \quad P_3 = (1-r)^2
        \vspace{-2mm}
    \end{equation}
\end{small}

A reconstructed sample is presented in \cref{fig:mae2}, with the masking ratio during inference set to 0.25. The results indicate that regions corresponding to all three masking cases are reconstructed with high quality.

\subsubsection{Dynamic Masking Ratio for Global Extraction}
\begin{figure}[H]
  \centering
  \begin{subfigure}{0.32\linewidth}
  \centering
  \includegraphics[width=0.72in]{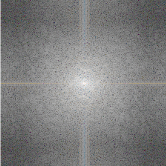}
    \caption{input}
    \label{fig:mae_origin3}
  \end{subfigure}
    \begin{subfigure}{0.32\linewidth}
  \centering
\includegraphics[width=0.72in]{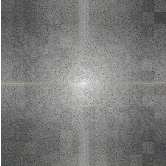}
    \caption{reconstructed}
    \label{fig:mae_recon3}
  \end{subfigure}
    \begin{subfigure}{0.32\linewidth}
  \centering
\includegraphics[width=0.72in]{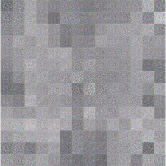}
    \caption{difference}
    \label{fig:mae_diff3}
  \end{subfigure}
  \vspace{-2mm}
  \caption{During training, adopt our loss function but set $r$ as a fixed value 0.3. During inference, using mask with ratio of 0.}
  \label{fig:mae3}
  \vspace{-2mm}
\end{figure}

Although the reconstructed sample in \cref{fig:mae2} demonstrates high-quality reconstruction with a masking ratio of 0.25 during inference, it can be observed that the global magnitude of the spectrum is not perfectly recovered as shown in \cref{fig:mae3}: when the model is trained with the same settings but evaluated with a mask ratio of 0, the reconstructed magnitude of the spectrum (\cref{fig:mae_recon3}) exhibits inconsistencies between blocks. Additionally, we find that if the mask ratio for inference significantly varies from the mask ratio during training, the performance of spectral reconstruction can be negatively influenced.

This observation inspires us to introduce a dynamic masking mechanism during training, where the mask ratio is randomly varied across different batches. Specifically, we define three levels of masking: heavily masked, slightly masked, and not masked, with corresponding mask ratios \( r_1 \), \( r_2 \), and \( r_3 \) set to 0.3, 0.15, and 0.0, respectively. Within each mini-batch, the mask ratio is consistent (i.e., it is either \( r_1 \), \( r_2 \), or \( r_3 \)), but the specific mask ratio used varies between different batches.

\begin{figure*}[ht]
    \centering
    \includegraphics[width=6.2in]{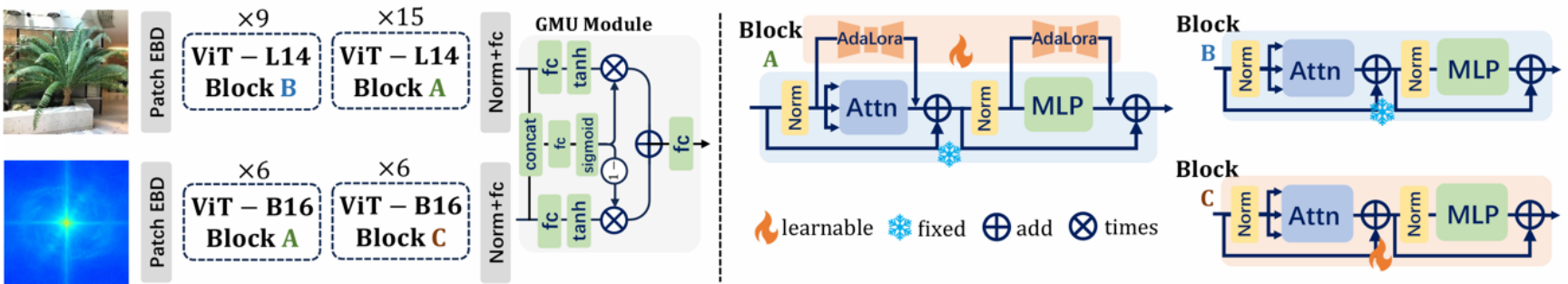}
    \vspace{-2mm}
    \caption{Spatial-frequency architecture showing the different blocks with distinct fine-tuning strategies across various network stages.}
    \label{fig:framework}
    \vspace{-4mm}
\end{figure*}

Different from \( \mathop{\mathcal{L}}_{r \neq 0}(X, X' | r)\) for $r_1$, $r_2$, when \(r_3 = 0\):

\begin{small}
\vspace{-4mm}
\begin{equation}
\mathcal{L}(X, X'|r_3=0) = \frac{1}{N^2} \sum_{i=0}^{N-1} \sum_{j=0}^{N-1} \mathcal{L}_{B(i,j)} 
\vspace{-2mm}
\end{equation}
\end{small}

In the experiments, it was observed that the order of magnitude for the loss function varies with different mask ratios. To mitigate potential instability in training caused by significant fluctuations in gradient updates across batches for varying \(r\) values, we introduce a scaling factor. Specifically, we compute the expected loss \(\mathbb{E}[\mathcal{L}(X, X'|r)]\) for each \(r\), and then scale the individual losses \(\mathcal{L}(X, X'|r_1)\), \(\mathcal{L}(X, X'|r_2)\), and \(\mathcal{L}(X, X'|r_3)\) by the reciprocal of their respective expectations. This normalization ensures a more consistent gradient descent process, thereby enhancing the stability of the neural network's training across different mask ratio configurations.

To compute the expectation of the reconstruction loss, we assume that \(\mathcal{L}_{B(i,j)}\) follows a \(\chi\) distribution and is independent of the scenario type \(t\). A detailed derivation proving that \(\mathcal{L}_{B(i,j)}\) conforms to a \(\chi\) distribution is provided in the Appendix A. Our goal is to determine \(\mathbb{E}[\mathcal{L}(X, X'|r_k)]\) for \(k = 1, 2, 3\), where \(\mathbb{E}[\cdot]\) represents the expectation over the specified distribution.

For $k = 1$ and $k = 2$, the $r_k \neq 0$, we acquire:

\begin{small}
\vspace{-3mm}
    \begin{equation}
     \mathbb{E}(\mathcal{L}(X, X'|r_k)) = \sum_{t=1}^3 P_t \mathbb{E}[\mathcal{L}(X, X'|r_k)|t]   
     \vspace{-1mm}
    \end{equation}
\end{small}

For each $t$, the $\mathbb{E}[\mathcal{L}(X, X'|r_k)|t]$ is equal to:
\begin{small}
\vspace{-2mm}
\begin{equation}
\begin{aligned}
 & -\frac{1}{N^2} \sum_{i=0}^{N-1} \sum_{j=0}^{N-1} \alpha_t \mathbb{E}[(1 - \mathcal{L}_{B(i,j)})^\gamma \log \mathcal{L}_{B(i,j)}] \\
 & = -\alpha_t \mathbb{E}[(1 - \mathcal{L}_B)^\gamma \log \mathcal{L}_B]
 \end{aligned}
 \vspace{-2mm}
\end{equation}
\end{small}

Therefore, for $k = 1$ and $k = 2$,
\begin{small}
\vspace{-2mm}
\begin{equation}
    \begin{aligned}
        & \mathbb{E}(\mathcal{L}(X, X'|r_k)) = - \left( \sum_{t=1}^3 P_t \alpha_t \right) \mathbb{E}[(1 - \mathcal{L}_B)^\gamma \log \mathcal{L}_B] \\
        & = - \frac{3 r_k^2 (1-r_k)^2 }{3r_k^2-3r_k+2} \mathbb{E}[(1 - \mathcal{L}_B)^\gamma \log \mathcal{L}_B]
    \end{aligned}
    \vspace{-2mm}
\end{equation}
\end{small}
where $\mathbb{E}[(1 - \mathcal{L}_B)^\gamma \log \mathcal{L}_B]$ is equal to:

\begin{small}
\vspace{-2mm}
\begin{equation}
\int_{0}^{\infty} (1 - x)^\gamma \log x \cdot \frac{1}{2} e^{\frac{-(x+\lambda)}{2}} \left(\frac{x}{\lambda}\right)^{\frac{k}{4}-\frac{1}{2}} I_{k/2-1}(\sqrt{\lambda x})  \mathrm{d} x
\vspace{-2mm}
\end{equation}
\end{small}
where $I_\nu(z)$ is the modified Bessel function of the first kind of order $\nu$.
Detailed steps are provided in the Appendix A.

For $k = 3$, we can easily get:
\begin{small}
\vspace{-2mm}
\begin{equation}
\mathbb{E}[\mathcal{L}(X, X'|r_3)] = \mathbb{E} \left[ \frac{1}{N^2} \sum_{i=0}^{N-1} \sum_{j=0}^{N-1} \mathcal{L}_{B(i,j)} \right] = \mathbb{E}[\mathcal{L}_B]
\vspace{-2mm}
\end{equation}
\end{small}

Therefore, for mask ratio of to 0.3, 0.15, we have the scaled loss function \(\mathcal{L}_k\) as:
\begin{small}
\vspace{-2mm}
\begin{equation}
\frac{1}{\mathbb{E}(\mathcal{L}(X, X'|r_k))} \cdot \mathcal{L}(X, X'|r_k)
\vspace{-2mm}
\end{equation}
\end{small}

\begin{figure}[H]
  \centering
  \begin{subfigure}{0.24\linewidth}
  \centering
  \includegraphics[width=0.72in]{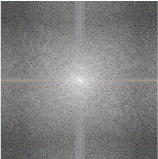}
    \caption{input}
    \label{fig:mae_origin4}
  \end{subfigure}
  \begin{subfigure}{0.24\linewidth}
  \centering
\includegraphics[width=0.72in]{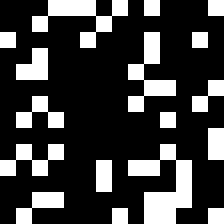}
    \caption{mask for (c)}
    \label{fig:mae_mask4}
  \end{subfigure}
    \begin{subfigure}{0.24\linewidth}
  \centering
\includegraphics[width=0.72in]{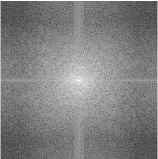}
    \caption{from (a), (b)}
    \label{fig:mae_recon4_25mask}
  \end{subfigure}
    \begin{subfigure}{0.24\linewidth}
  \centering
\includegraphics[width=0.72in]{figures/MAE/recon4_25mask.png}
    \caption{from (a) only}
    \label{fig:mae_recon4_0mask}
  \end{subfigure}
  \vspace{-2mm}
  \caption{Dynamically set $r=0.3, 0.15, 0$. (a) original magnitude, (b) mask with ratio of 0.25, (c) the reconstructed magnitude using (b) mask, (d) the reconstructed magnitude without mask.}
  \vspace{-2mm}
  \label{fig:mae4}
\end{figure}

We show from experimental results that it is necessary to introduce dynamic masking ratios to capture the global reconstruction during pre-training of FFiT. After dynamically setting $r$ to $r_1$, $r_2$, $r_3$ $=0.3,0.15,0$ respectively, the global reconstruction is perfect which can be observed from \cref{fig:mae4}.

\subsection{Design of Architecture}
We employ a Transformer-based backbone consisting of ViT-L14 blocks for the spatial branch and ViT-B16 blocks for the frequency branch. This configuration is based on experimental findings that indicate increasing the complexity of the frequency branch does not substantially enhance performance but instead introduces greater instability during training and reduces inference speed. The AdaLoRA \cite{adalora} blocks are introduced in both spatial and spectral branches to dynamically fine-tune the pre-trained parameters.
We utilize Gated-Multimodal-Unit (GMU) \cite{arevalo2017gated} for information fusion. The sequence of blocks are visualized in \cref{fig:framework}.

%% file: Sections/theory.tex
\vspace{-2mm}
\section{Quantitative Analysis of Generalization Ability for Fake Detectors}
\label{sec:theory_analysis}
\begin{figure}[ht]
    \centering
    \includegraphics[width=0.95\linewidth]{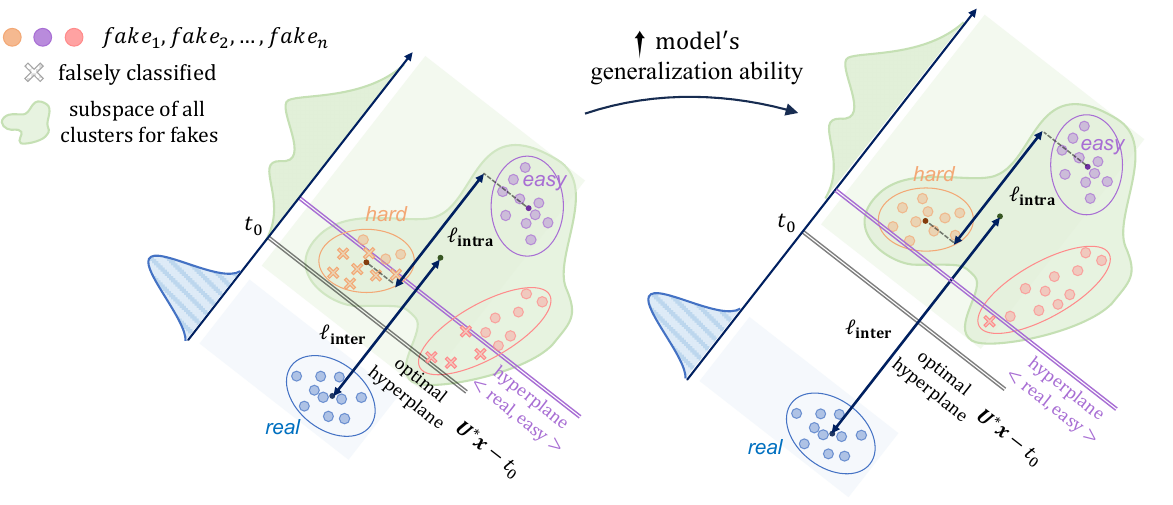}
    \vspace{-2mm}
    \caption{As the model's generalization capacity strengthens, the inter-cluster distance between real and fake data increases, leading to fewer misclassifications.}
    \label{fig:theory}
    \vspace{-2mm}
\end{figure}

In the numerical experimentation, it is observed that the performance of fake detectors trained on fake images generated by DM, NeRF, and 3DGS is generally better than that of detectors trained on GANs. This observation aligns with the findings in \cite{ojha2023towards,ricker2022towards}, which suggest that fake detectors trained on \textit{difficult} fake samples with less domain-specific information typically perform better. Additionally, we observe that this performance gap diminishes as the representation ability of the backbone architecture increases. To provide a quantitative analysis of this phenomenon, we examine the distribution of clustering features. The experimental observations indicate that as the representation capability of the backbone architecture improves, the distances along the direction perpendicular to the optimal hyperplane for real/fake classification between fake-fake clusters become relatively small compared to the distances between real-fake clusters. This explains why the generalization ability of larger models is less sensitive to whether the detector is trained on easy or difficult fake samples. An illustration of this phenomenon is provided in \cref{fig:theory}. The figure shows that as the model's generalization capability to distinguish between the real and fake images improves, there is a notable increase in the inter-cluster distance between the fake and real feature spaces, while the subspace of all fakes remains relatively stable. This observation explains why the number of incorrectly classified samples decreases for models trained on real and easily distinguishable fake data.

It is assumed that the parameters of the backbone are fixed to extract the deep features of an image as $f_i\in \mathbb{R}^d$, we assign the labels of real data as 0 and the labels of fake data as 1, thus, the label $l_i \in \{0,1\}$. The features are multiplied by $\boldsymbol{U}\in \mathbb{R}^{d\times1}$ to give the projected scores in one dimension. We consider the challenging case that unseen fake distribution does not fit well enough to the detector due to outlier data points always occurring beyond the cluster envelope, while the predicted results for real data are stable. We denote the error \( e_i \) of each data point projected by the hyperplane during training:

\begin{small}
\vspace{-2mm}
    \begin{equation}
    e_i = l_i - z_i \theta_i - f_i^\top \boldsymbol{U}
    \vspace{-2mm}
\end{equation}
\end{small}
where $z_i = 0$ if real and $z_i = 1$ if fake. An explicit data-dependent variable \( \theta_i \) is introduced to model this small amount of large error.

The item \( l_i - z_i \theta_i \) can be regarded as soft label of unseen fake data to compute the optimal hyperplane bound without the negative influence from sparse outlier data points.
Based on this model, we aim to acquire the optimal $\boldsymbol{U}$ for the best separation on the testing dataset as:

\begin{figure}[b]
  \centering
  \begin{subfigure}{0.32\linewidth}
  \centering
  \includegraphics[width=0.74in]{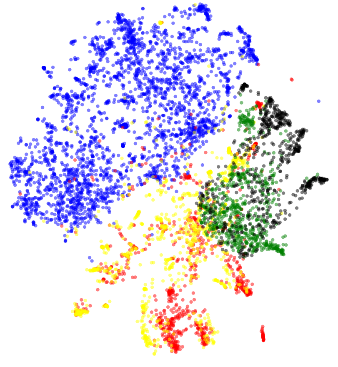}
    \caption{CNNSpot \cite{wang2020cnn}}
    \label{fig:distribute1}
  \end{subfigure}
    \begin{subfigure}{0.32\linewidth}
  \centering
\includegraphics[width=0.8in]{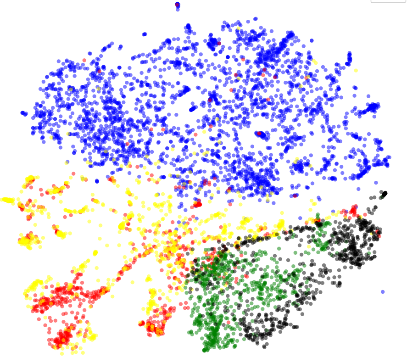}
    \caption{UniFD \cite{ojha2023towards}}
    \label{fig:distribute2}
  \end{subfigure}
    \begin{subfigure}{0.32\linewidth}
  \centering
\includegraphics[width=0.8in]{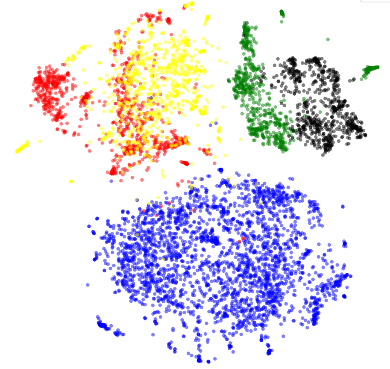}
    \caption{multimodal (\textbf{Ours})}
    \label{fig:distribute3}
  \end{subfigure}
  \vspace{-2mm}
  \caption{Samples of Distributions projected to two dimensions.}
  \label{fig:distribute}
  \vspace{-2mm}
\end{figure}

\begin{small}
    \vspace{-2mm}
    \begin{equation}
    \hspace{-2mm}
             \boldsymbol{U}^*, \boldsymbol{\Theta}^* = \mathop{\mathrm{argmin}}_{\boldsymbol{U}, \boldsymbol{\Theta}} 
         \sum_{i = 0}^{n} \left[ \frac{1}{2}(l_i - z_i \theta_i - f_i^\top \boldsymbol{U})^2 + \lambda |z_i \theta_i| \right]
         \vspace{-2mm}
\end{equation}
\end{small}
where $l_i$, $f_i$, $z_i$, $\theta_i$ are stacked as $\boldsymbol{L}$, $\boldsymbol{F}$, $\boldsymbol{Z}$, $\boldsymbol{\Theta}$, respectively.
Given the \( \ell_1 \) penalty imposed on \( \boldsymbol{\Theta} \) to encourage the sparse solution, we can directly compute the \( \boldsymbol{U}^* \) after \( \boldsymbol{\Theta}^* \) is resolved as follows:

\begin{small}
    \vspace{-2mm}
    \begin{equation}
 \boldsymbol{U}^* = (\boldsymbol{F}^\top \boldsymbol{F})^\dagger \boldsymbol{F}^\top (\boldsymbol{L} - \boldsymbol{Z} \otimes \boldsymbol{\Theta}^*)
 \vspace{-2mm}
\end{equation}
\end{small}
where \( \dagger \) denotes Moore-Penrose inverse, \( \otimes \) denotes Kronecker product.
Following \cite{fan2024test}, we define \( \widetilde{\boldsymbol{F}} = \boldsymbol{I} - \boldsymbol{F}(\boldsymbol{F}^\top \boldsymbol{F})^\dagger \boldsymbol{F}^\top \) and \( \widetilde{\boldsymbol{L}} = \widetilde{\boldsymbol{F}} \boldsymbol{L} \), we simplify the objective as:

\begin{small}
\vspace{-2mm}
    \begin{equation}
    \boldsymbol{\Theta}^* = \mathop{\mathrm{argmin}}_{\boldsymbol{\Theta}} \frac{1}{2} ||\widetilde{\boldsymbol{L}} - \widetilde{\boldsymbol{F}} (\boldsymbol{Z} \otimes \boldsymbol{\Theta})||_2^2 + \lambda ||\boldsymbol{Z} \otimes \boldsymbol{\Theta}||_1
    \vspace{-2mm}
\end{equation}
\end{small}

After selecting the proper $\lambda$ to resolve the \( \boldsymbol{\Theta}^* \) using Powell optimization algorithm \cite{powell1964efficient}, we select those rows of $\boldsymbol{F}$ and $\boldsymbol{L}$ corresponding to the top 80\% lowest values of $\boldsymbol{Z} \otimes \boldsymbol{\Theta}$ to acquire \( \boldsymbol{U}^*\).

\begin{small}
\vspace{-3mm}
    \begin{equation}
        \boldsymbol{U}^* = (\boldsymbol{F}_{\mathrm{low}}^\top \boldsymbol{F}_{\mathrm{low}})^\dagger \boldsymbol{F}_{\mathrm{low}}^\top \boldsymbol{L}_{\mathrm{low}}
        \vspace{-2mm}
    \end{equation}
\end{small}

It is assumed that the best threshold of scores in 1D for separation is $t_0$, which is set to $0.5$. Consequently, the optimal high-dimensional hyperplane in feature space is expressed as:
\begin{small}
\vspace{-2mm}
    \begin{equation}
    \boldsymbol{U}^*f_i - t_0=0
    \vspace{-1mm}
\end{equation}
\end{small}

For a point $x_i$ in high dimension, the distance $d$ to the hyperplane is represented as:

\begin{small}
\vspace{-3mm}
    \begin{equation}
    d= |\boldsymbol{U}^*x_i-t_0| / ||\boldsymbol{U}^*||
    \vspace{-1mm}
\end{equation}
\end{small}

We develop the index \(\rho\) to estimate the ratio of the distances between the most difficult and easiest fake clusters to the distances between real and fake clusters, measured along the direction perpendicular to the optimal hyperplane.

\begin{small}
\vspace{-4mm}
\begin{equation}
\hspace{-3mm}
    \rho = \frac{\mathop{\mathrm{max}}\limits_{m\neq n}\left( \sum\limits_{x_i\in G_m} \frac{|\boldsymbol{U}^*x_i-t_0|}{||\boldsymbol{U}^*||} - \sum\limits_{x_i\in G_n} \frac{|\boldsymbol{U}^*x_i-t_0|}{||\boldsymbol{U}^*||}\right)}{\frac{1}{\mid \{ i\mid l_i=0 \} \mid} \sum\limits_{i\mid l_i=0} \frac{|\boldsymbol{U}^*x_i-t_0|}{||\boldsymbol{U}^*||} + \frac{1}{\mid \{ i\mid l_i=1 \} \mid} \sum\limits_{i\mid l_i=1} \frac{|\boldsymbol{U}^*x_i-t_0|}{||\boldsymbol{U}^*||}}
    \vspace{-2mm}
\end{equation}
\end{small}
where $G$ denotes the cluster of fake images during training.

The quantitative results of $\rho$ for different fake feature extraction methods are provided in \cref{sec:results_theory}.

%% file: Sections/datasets.tex
\vspace{-2mm}
\section{Datasets and Protocols}
We now present the experimental setup employed for evaluating the developed method for fake image detection.

\input{Tables/protocol_test}

\vspace{-2mm}
\subsection{Introduction to Our Database}
We have compiled 139 groups of consecutive 2D photos, each accompanied by the corresponding camera poses. These camera poses were either recorded during the photo capture process or calibrated using structure-from-motion techniques. To generate \textit{fake} images, we employed a variety of methods capable of producing realistic 3D representations from the aforementioned data. Specifically, we utilized different NeRF-based and 3DGS-based methods, labeled from \textbf{\uppercase\expandafter{\romannumeral1}} to \textbf{\uppercase\expandafter{\romannumeral8}}, to generate 3D scene representations. In \cref{tab:generated_db}, we highlight the number of successfully reconstructed 3D scenes generated by each method in blue. Once the 3D scenes are reconstructed, they are projected back to 2D images using the same projection parameters as those estimated from the original 2D photos. These rendered 2D images are considered \textit{fake}, and the numbers of the fake projected images are highlighted in red. In experiments, the original 2D photos used for reconstruction are regarded as \textit{real} images.

\subsection{Protocols for Training and Evaluation}
We use four types of training datasets ($\mathcal{A}, \mathcal{B}, \mathcal{C}, \mathcal{D}$), containing real images and fake images respectively generated by GAN, DM, NeRF, and 3DGS to train different fake detectors in the experiments. For the details of those training datasets, please refer to Appendix C.

For evaluation, we assess the performance on the fake images across several categories: those generated by NeRF or 3DGS, combinations of traditional generative methods with neural rendering, editable neural rendering, digital avatars (both heads and full bodies), and synthetic video frames with realistic 3D representations. These categories are systematically evaluated from group 1 to group 11.

We provide the details of the protocol for training the architecture in Appendix C.

\vspace{-2mm}

\input{Tables/main_results}

\input{Tables/results_unifd}

% \begin{table}[ht]
% \centering
% \tabcolsep=0.35cm
% \caption{Training and Evaluation Protocols}
% \label{tab:summarize_protocols}
% \fontsize{7pt}{8pt}\selectfont %
% \begin{tabular}{l|cccc}
% \noalign{{\color{black}\hrule height 1pt}}
% \diagbox{test}{train} & \begin{tabular}[c]{@{}c@{}}GAN\\ ($\mathcal{A}$)\end{tabular} & \begin{tabular}[c]{@{}c@{}}DM\\ ($\mathcal{B}$)\end{tabular} & \begin{tabular}[c]{@{}c@{}}NeRF\\ ($\mathcal{C}$)\end{tabular} & \begin{tabular}[c]{@{}c@{}}3DGS\\ ($\mathcal{D}$)\end{tabular} \\ \hline
% 1. \textbf{\uppercase\expandafter{\romannumeral1}} $\sim$ \textbf{\uppercase\expandafter{\romannumeral5}} & PT-$\mathcal{A}_1$ & PT-$\mathcal{B}_1$ & PT-\textcolor{red}{$\text{NN}$} & PT-$\mathcal{D}_1$ \\
% 2. \textbf{\uppercase\expandafter{\romannumeral6}} $\sim$ \textbf{\uppercase\expandafter{\romannumeral8}} & PT-$\mathcal{A}_2$ & PT-$\mathcal{B}_2$ & PT-$\mathcal{C}_2$ & PT-\textcolor{red}{$\text{GG}$} \\
% 3. Pix2NeRF & PT-$\mathcal{A}_3$ & PT-$\mathcal{B}_3$ & PT-$\mathcal{C}_3$ & PT-$\mathcal{D}_3$ \\
% \rotatebox{90}{...} & \rotatebox{90}{...} & \rotatebox{90}{...} & \rotatebox{90}{...} & \rotatebox{90}{...} \\
% 11. Sora & PT-$\mathcal{A}_{11}$ & PT-$\mathcal{B}_{11}$ & PT-$\mathcal{C}_{11}$ & PT-$\mathcal{D}_{11}$ \\ \noalign{{\color{black}\hrule height 1pt}}
% \end{tabular}
% \end{table}

%% file: Tables/protocol_test.tex
\begin{minipage}{\textwidth}
\hspace{-10mm}
\begin{minipage}[t]{0.2\textwidth}
\makeatletter\def\@captype{table}
\tabcolsep=0.04cm
\fontsize{7pt}{7.6pt}\selectfont
\centering
\caption{Collected data}
\vspace{-2mm}
\begin{tabular}{cc}
\toprule
\multicolumn{1}{c|}{\textbf{\uppercase\expandafter{\romannumeral1}} \cite{ingp}} & \textbf{\uppercase\expandafter{\romannumeral2}} \cite{tensorf} \\
\multicolumn{1}{c|}{i-ngp} & tensorf \\
\multicolumn{1}{c|}{\textcolor{blue}{115}/\textcolor{red}{52,927}} & \textcolor{blue}{109}/\textcolor{red}{49,475} \\ \hline
\multicolumn{1}{c|}{\textbf{\uppercase\expandafter{\romannumeral3}} \cite{tancik2023nerfstudio}} & \textbf{\uppercase\expandafter{\romannumeral4}} \cite{seathru} \\
\multicolumn{1}{c|}{nerfacto} & seaThru \\
\multicolumn{1}{c|}{\textcolor{blue}{122}/\textcolor{red}{70,073}} & \textcolor{blue}{97}/\textcolor{red}{34,557} \\ \hline
\multicolumn{1}{c|}{\textbf{\uppercase\expandafter{\romannumeral5}} \cite{pynerf}} & \textbf{\uppercase\expandafter{\romannumeral6}} \cite{3dgs} \\
\multicolumn{1}{c|}{pynerf} & 3DGS \\
\multicolumn{1}{c|}{\textcolor{blue}{84}/\textcolor{red}{27,457}} & \textcolor{blue}{28}/\textcolor{red}{3,661} \\ \hline
\multicolumn{1}{c|}{\textbf{\uppercase\expandafter{\romannumeral7}} \cite{tancik2023nerfstudio}} & \textbf{\uppercase\expandafter{\romannumeral8}} \cite{c3dgs} \\
\multicolumn{1}{c|}{splatfacto} & C-3DGS \\
\multicolumn{1}{c|}{\textcolor{blue}{115}/\textcolor{red}{43,908}} & \textcolor{blue}{36}/\textcolor{red}{14,446} \\ \bottomrule
\end{tabular}
\label{tab:generated_db}
\end{minipage}
\hspace{-5mm}
\begin{minipage}[t]{0.3\textwidth}
\makeatletter\def\@captype{table}
\tabcolsep=0.03cm
\fontsize{7pt}{8pt}\selectfont
\centering
\caption{Protocol of test data}
\vspace{-2mm}
\begin{tabular}{lcc}
\noalign{{\color{black}\hrule height 1pt}}
\multicolumn{1}{c}{Test group} & real/fake & Type \\ \hline
1 \textbf{\uppercase\expandafter{\romannumeral1}} $\sim$ \textbf{\uppercase\expandafter{\romannumeral5}} & 3,726/13,942
 & unseen NeRF \\
2 \textbf{\uppercase\expandafter{\romannumeral6}} $\sim$ \textbf{\uppercase\expandafter{\romannumeral8}} & 3,726/10,496
 & unseen 3DGS \\
3 Pix2NeRF \cite{pix2nerf} & 70,000/96,000 & GAN+NeRF \\
4 SketchFaceNeRF \cite{sketchfacenerf} & 70,000/90,000 & GAN+NeRF \\
5 DreamFusion \cite{dreamfusion} & 10,000/10,600 & DM+NeRF \\
6 GSGEN \cite{gsgen} & 10,000/9,540 & DM+3DGS \\
7 instruct-N2N \cite{instruct_n2n} & 3,174/40,559 & editable NeRF \\
8 instruct-GS2GS \cite{instruct_gs2gs} & 3,174/40,559 & editable 3DGS \\
9 GeneFace++ \cite{genefacepp} & 9,100/9,100 & Avatar NeRF \\
10 SplattingAvatar \cite{splattingavatar} & 33,728/33,715 & Avatar 3DGS \\
11 Sora \cite{sora} & 60,000/60,531 & 3D-realistic video \\ \noalign{{\color{black}\hrule height 1pt}}
\end{tabular}
\label{tab:methods_mapping}
\end{minipage}
\end{minipage}

%% file: Tables/main_results.tex
\begin{table*}[ht]
\centering
\setlength{\arrayrulewidth}{0.05pt}
\tabcolsep=0.072cm
\caption{Comparative results of different methods in \%. }
\vspace{-2mm}
\label{tab:performance_ap_auroc}
\fontsize{6.2pt}{6.7pt}\selectfont %
\begin{tabular}{c|cccccccccccccc|cccccccccccc}
 \noalign{{\color{black}\hrule height 1pt}}
 &  & \multirow{2}{*}{group} & \multicolumn{11}{c}{Average Precision} &  & \multicolumn{11}{c}{AUROC} &  \\
 &  &  & \textbf{1} & \textbf{2} & \textbf{3} & \textbf{4} & \textbf{5} & \textbf{6} & \textbf{7} & \textbf{8} & \textbf{9} & \textbf{10} & \textbf{11} & ave & \textbf{1} & \textbf{2} & \textbf{3} & \textbf{4} & \textbf{5} & \textbf{6} & \textbf{7} & \textbf{8} & \textbf{9} & \textbf{10} & \textbf{11} & ave \\ \hline
\multirow{20}{*}{\rotatebox{90}{spatial- based}} & \multirow{5}{*}{\begin{tabular}[c]{@{}c@{}}CNNSpot\\ \cite{wang2020cnn} \end{tabular}} & $\mathcal{A}$ & 83.20 & 76.19 & 89.75 & 95.66 & 67.04 & 56.14 & 98.11 & 98.73 & 65.71 & 68.33 & 43.45 & \cellcolor{lightgray}76.57 & 56.02 & 52.16 & 91.47 & 95.04 & 69.66 & 64.34 & 78.60 & 84.76 & 65.29 & 74.25 & 39.21 & \cellcolor{lightgray}70.07 \\
 &  & $\mathcal{B}$ & 88.03 & 75.21 & 88.49 & 97.29 & 84.20 & 93.21 & 98.05 & 98.49 & 55.36 & 49.08 & 60.45 & \cellcolor{lightgray}80.71 & 66.04 & 52.75 & 88.39 & 96.94 & 86.78 & 94.68 & 81.07 & 85.27 & 60.69 & 43.59 & 60.62 & \cellcolor{lightgray}74.26 \\
 &  & $\mathcal{C}$ & / & 87.06 & 72.89 & 80.47 & 70.36 & 82.55 & 99.97 & 99.79 & 58.34 & 84.62 & 67.21 & \cellcolor{lightgray}80.33 & / & 66.67 & 73.15 & 79.92 & 64.83 & 79.33 & 99.65 & 97.61 & 59.31 & 92.05 & 68.19 & \cellcolor{lightgray}78.07 \\
 &  & $\mathcal{D}$ & 94.31 & / & 92.80 & 65.68 & 61.84 & 96.44 & 99.36 & 99.95 & 65.43 & 73.96 & 59.06 & \cellcolor{lightgray}80.88 & 80.77 & / & 92.53 & 64.92 & 59.24 & 94.46 & 92.74 & 99.46 & 62.12 & 84.99 & 55.47 & \cellcolor{lightgray}78.67 \\
 &  & \cellcolor{lightgray}ave & \cellcolor{lightgray}88.51 & \cellcolor{lightgray}79.49 & \cellcolor{lightgray}85.98 & \cellcolor{lightgray}84.77 & \cellcolor{lightgray}70.86 & \cellcolor{lightgray}82.08 & \cellcolor{lightgray}98.87 & \cellcolor{lightgray}99.24 & \cellcolor{lightgray}61.21 & \cellcolor{lightgray}68.99 & \cellcolor{lightgray}57.54 &  & \cellcolor{lightgray}67.61 & \cellcolor{lightgray}57.19 & \cellcolor{lightgray}86.38 & \cellcolor{lightgray}84.20 & \cellcolor{lightgray}70.12 & \cellcolor{lightgray}83.20 & \cellcolor{lightgray}88.01 & \cellcolor{lightgray}91.77 & \cellcolor{lightgray}61.85 & \cellcolor{lightgray}73.72 & \cellcolor{lightgray}55.87 &  \\ 
 \cline{2-27}
 & \multirow{5}{*}{\begin{tabular}[c]{@{}c@{}}UniFD \\ \cite{ojha2023towards}\end{tabular}} & $\mathcal{A}$ & 93.82 & 77.02 & 92.78 & 92.46 & 61.63 & 73.15 & 96.09 & 97.72 & 60.05 & 61.19 & 42.40 & \cellcolor{lightgray}77.12 & 80.56 & 55.90 & 91.39 & 91.44 & 61.55 & 77.40 & 72.49 & 84.41 & 62.76 & 57.79 & 37.01 & \cellcolor{lightgray}70.25 \\
 &  & $\mathcal{B}$ & 94.19 & 88.26 & 76.57 & 99.31 & 93.66 & 93.07 & 99.39 & 99.79 & 64.62 & 71.77 & 48.84 & \cellcolor{lightgray}84.50 & 82.21 & 73.77 & 67.68 & 99.03 & 92.94 & 93.55 & 94.00 & 97.97 & 67.69 & 70.60 & 43.21 & \cellcolor{lightgray}80.24 \\
 &  & $\mathcal{C}$ & / & 83.65 & 75.23 & 82.14 & 76.50 & 66.72 & 99.89 & 99.91 & 60.41 & 72.55 & 71.51 & \cellcolor{lightgray}78.85 & / & 65.56 & 71.17 & 81.80 & 71.98 & 73.78 & 98.83 & 99.02 & 57.84 & 70.60 & 72.38 & \cellcolor{lightgray}76.30 \\
 &  & $\mathcal{D}$ & 95.67 & / & 92.09 & 86.97 & 73.83 & 62.63 & 99.82 & 99.97 & 64.46 & 83.61 & 82.42 & \cellcolor{lightgray}84.15 & 86.02 & / & 90.08 & 85.73 & 71.04 & 67.37 & 97.92 & 99.72 & 64.04 & 82.91 & 82.89 & \cellcolor{lightgray}82.77 \\
 &  & \cellcolor{lightgray}ave & \cellcolor{lightgray}94.56 & \cellcolor{lightgray}82.97 & \cellcolor{lightgray}84.16 & \cellcolor{lightgray}90.22 & \cellcolor{lightgray}76.40 & \cellcolor{lightgray}73.89 & \cellcolor{lightgray}98.79 & \cellcolor{lightgray}99.34 & \cellcolor{lightgray}62.38 & \cellcolor{lightgray}72.28 & \cellcolor{lightgray}61.29 &  & \cellcolor{lightgray}82.93 & \cellcolor{lightgray}65.07 & \cellcolor{lightgray}80.08 & \cellcolor{lightgray}89.50 & \cellcolor{lightgray}74.37 & \cellcolor{lightgray}78.02 & \cellcolor{lightgray}90.81 & \cellcolor{lightgray}95.28 & \cellcolor{lightgray}63.08 & \cellcolor{lightgray}70.47 & \cellcolor{lightgray}58.87 &  \\ \cline{2-27}
 & \multirow{5}{*}{\begin{tabular}[c]{@{}c@{}}MoeFD\\ \cite{moe_lora}\end{tabular}} & $\mathcal{A}$ & 86.08 & 82.50 & 91.43 & 98.72 & 73.71 & 85.68 & 96.32 & 99.78 & 64.22 & 60.42 & 62.28 & \cellcolor{lightgray}81.92 & 60.41 & 63.85 & 90.86 & 98.42 & 74.50 & 86.10 & 64.49 & 97.36 & 58.06 & 60.16 & 59.42 & \cellcolor{lightgray}73.97 \\
 &  & $\mathcal{B}$ & 87.93 & 85.03 & 88.65 & 99.85 & 98.94 & 94.94 & 98.22 & 99.26 & 66.02 & 65.32 & 64.73 & \cellcolor{lightgray}86.26 & 65.73 & 69.70 & 88.80 & 99.80 & 98.82 & 95.05 & 81.87 & 91.62 & 60.29 & 63.93 & 62.25 & \cellcolor{lightgray}79.81 \\
 &  & $\mathcal{C}$ & / & 87.32 & 91.20 & 88.96 & 82.23 & 78.09 & 96.77 & 99.82 & 61.19 & 67.69 & 69.58 & \cellcolor{lightgray}82.29 & / & 74.94 & 90.66 & 89.34 & 82.87 & 78.76 & 68.28 & 97.86 & 54.36 & 65.77 & 67.71 & \cellcolor{lightgray}77.06 \\
 &  & $\mathcal{D}$ & 89.68 & / & 90.13 & 90.46 & 78.20 & 88.69 & 98.13 & 99.49 & 69.00 & 71.25 & 74.23 & \cellcolor{lightgray}84.93 & 70.94 & / & 89.86 & 90.63 & 78.94 & 89.01 & 80.97 & 94.11 & 63.99 & 68.62 & 72.81 & \cellcolor{lightgray}79.99 \\
 &  & \cellcolor{lightgray}ave & \cellcolor{lightgray}87.89 & \cellcolor{lightgray}84.95 & \cellcolor{lightgray}90.35 & \cellcolor{lightgray}94.49 & \cellcolor{lightgray}83.27 & \cellcolor{lightgray}86.85 & \cellcolor{lightgray}97.36 & \cellcolor{lightgray}99.58 & \cellcolor{lightgray}65.10 & \cellcolor{lightgray}66.17 & \cellcolor{lightgray}67.70 &  & \cellcolor{lightgray}65.69 & \cellcolor{lightgray}69.49 & \cellcolor{lightgray}90.04 & \cellcolor{lightgray}94.54 & \cellcolor{lightgray}83.78 & \cellcolor{lightgray}87.23 & \cellcolor{lightgray}73.90 & \cellcolor{lightgray}95.23 & \cellcolor{lightgray}59.17 & \cellcolor{lightgray}64.62 & \cellcolor{lightgray}65.54 &  \\ \cline{2-27}
 & \multirow{5}{*}{\begin{tabular}[c]{@{}c@{}}RGB branch\\ (ours)\end{tabular}} & $\mathcal{A}$ & 93.38 & 92.02 & 78.57 & 91.25 & 98.95 & 98.12 & 97.76 & 97.98 & 63.08 & 49.43 & 62.18 & \cellcolor{lightgray}83.88 & 85.38 & 85.15 & 76.32 & 93.47 & 99.20 & 98.52 & 83.25 & 83.53 & 63.80 & 54.81 & 64.16 & \cellcolor{lightgray}80.69 \\
 &  & $\mathcal{B}$ & 87.67 & 83.77 & 90.95 & 98.51 & 99.54 & 99.14 & 98.35 & 99.81 & 66.78 & 59.04 & 56.05 & \cellcolor{lightgray}85.42 & 68.88 & 68.17 & 89.52 & 98.37 & 99.60 & 99.27 & 84.12 & 97.76 & 65.19 & 57.85 & 53.09 & \cellcolor{lightgray}80.17 \\
 &  & $\mathcal{C}$ & / & 99.25 & 86.01 & 94.57 & 99.33 & 98.16 & 99.83 & 99.87 & 65.18 & 53.31 & 54.88 & \cellcolor{lightgray}85.04 & / & 98.79 & 84.82 & 95.95 & 99.49 & 98.62 & 98.70 & 98.99 & 66.91 & 58.71 & 54.19 & \cellcolor{lightgray}85.52 \\
 &  & $\mathcal{D}$ & 99.45 & / & 92.99 & 99.76 & 99.94 & 99.62 & 99.98 & 99.99 & 77.79 & 77.88 & 57.42 & \cellcolor{lightgray}90.48 & 98.05 & / & 90.76 & 99.71 & 99.94 & 99.68 & 99.84 & 99.96 & 74.64 & 76.06 & 56.12 & \cellcolor{lightgray}89.48 \\
 &  & \cellcolor{lightgray}ave & \cellcolor{lightgray}93.50 & \cellcolor{lightgray}91.68 & \cellcolor{lightgray}87.13 & \cellcolor{lightgray}96.02 & \cellcolor{lightgray}\textbf{99.44} & \cellcolor{lightgray}98.76 & \cellcolor{lightgray}98.98 & \cellcolor{lightgray}99.41 & \cellcolor{lightgray}68.21 & \cellcolor{lightgray}59.92 & \cellcolor{lightgray}57.63 &  & \cellcolor{lightgray}84.10 & \cellcolor{lightgray}\textbf{84.04} & \cellcolor{lightgray}85.36 & \cellcolor{lightgray}96.88 & \cellcolor{lightgray}\textbf{99.56} & \cellcolor{lightgray}99.02 & \cellcolor{lightgray}91.48 & \cellcolor{lightgray}95.06 & \cellcolor{lightgray}\textbf{67.64} & \cellcolor{lightgray}61.86 & \cellcolor{lightgray}56.89 &  \\ \hline
\multirow{10}{*}{\rotatebox{90}{spectral- based}} & \multirow{5}{*}{\begin{tabular}[c]{@{}c@{}}Freq-spec\\ \cite{zhang2019detecting}\end{tabular}} & $\mathcal{A}$ & 81.83 & 74.59 & 94.94 & 79.52 & 79.19 & 77.15 & 96.64 & 97.69 & 53.44 & 67.29 & 46.69 & \cellcolor{lightgray}77.18 & 53.15 & 49.77 & 96.87 & 81.84 & 82.70 & 84.86 & 70.39 & 78.00 & 57.58 & 68.45 & 47.27 & \cellcolor{lightgray}70.08 \\
 &  & $\mathcal{B}$ & 83.15 & 76.44 & 84.47 & 79.26 & 80.74 & 76.85 & 96.58 & 96.88 & 56.93 & 70.78 & 57.35 & \cellcolor{lightgray}78.13 & 58.51 & 54.48 & 87.94 & 80.60 & 81.49 & 85.83 & 71.68 & 73.74 & 58.55 & 74.74 & 59.35 & \cellcolor{lightgray}71.54 \\
 &  & $\mathcal{C}$ & / & 87.48 & 97.07 & 90.64 & 59.34 & 92.39 & 99.06 & 99.05 & 57.92 & 64.83 & 76.10 & \cellcolor{lightgray}82.39 & / & 70.42 & 96.18 & 84.73 & 46.72 & 94.21 & 93.10 & 93.33 & 57.94 & 76.46 & 73.26 & \cellcolor{lightgray}78.64 \\
 &  & $\mathcal{D}$ & 96.60 & / & 89.56 & 81.52 & 54.01 & 90.47 & 99.71 & 99.68 & 60.43 & 86.79 & 72.76 & \cellcolor{lightgray}83.15 & 89.70 & / & 91.27 & 68.74 & 65.54 & 93.68 & 97.17 & 96.86 & 63.02 & 86.37 & 70.29 & \cellcolor{lightgray}82.26 \\
 &  & \cellcolor{lightgray}ave & \cellcolor{lightgray}87.19 & \cellcolor{lightgray}79.50 & \cellcolor{lightgray}91.51 & \cellcolor{lightgray}82.73 & \cellcolor {lightgray}68.32 & \cellcolor {lightgray}84.22 & \cellcolor {lightgray}98.00 & \cellcolor {lightgray}98.33 & \cellcolor {lightgray}57.18 & \cellcolor {lightgray}72.42 & \cellcolor {lightgray}63.23 &  & \cellcolor {lightgray}67.12 & \cellcolor {lightgray}58.22 & \cellcolor {lightgray}93.06 & \cellcolor {lightgray}78.98 & \cellcolor {lightgray}69.11 & \cellcolor {lightgray}89.64 & \cellcolor {lightgray}83.08 & \cellcolor {lightgray}85.48 & \cellcolor {lightgray}59.27 & \cellcolor {lightgray}76.51 & \cellcolor {lightgray}62.54 &  \\ \cline{2-27} 
 & \multirow{5}{*}{FFiT} & $\mathcal{A}$ & 89.55 & 82.41 & 99.96 & 96.19 & 97.75 & 97.19 & 97.78 & 98.08 & 63.52 & 81.10 & 67.64 & \cellcolor{lightgray}88.29 & 69.86 & 62.46 & 99.95 & 95.17 & 97.67 & 97.98 & 78.67 & 81.82 & 62.33 & 81.76 & 67.67 & \cellcolor{lightgray}81.39 \\
 &  & $\mathcal{B}$ & 97.67 & 90.98 & 95.54 & 98.73 & 90.30 & 93.95 & 99.00 & 99.12 & 64.16 & 55.15 & 84.97 & \cellcolor{lightgray}88.14 & 92.30 & 80.30 & 92.94 & 98.94 & 90.04 & 95.21 & 88.95 & 90.53 & 67.99 & 50.42 & 82.58 & \cellcolor{lightgray}84.56 \\
 &  & $\mathcal{C}$ & / & 82.66 & 97.90 & 99.91 & 78.53 & 72.24 & 97.67 & 97.59 & 51.10 & 75.82 & 87.31 & \cellcolor{lightgray}84.07 & / & 64.04 & 98.06 & 99.87 & 78.51 & 79.04 & 81.11 & 80.52 & 50.89 & 73.27 & 84.65 & \cellcolor{lightgray}79.00 \\
 &  & $\mathcal{D}$ & 92.75 & / & 99.11 & 94.38 & 64.26 & 96.77 & 99.47 & 99.66 & 50.43 & 96.48 & 80.78 & \cellcolor{lightgray}87.41 & 79.06 & / & 98.84 & 92.77 & 54.91 & 95.81 & 94.62 & 96.95 & 54.00 & 96.76 & 78.53 & \cellcolor{lightgray}84.23 \\
 &  & \cellcolor{lightgray}ave & \cellcolor{lightgray}93.32 & \cellcolor{lightgray}85.35 & \cellcolor{lightgray}98.13 & \cellcolor{lightgray}97.30 & \cellcolor{lightgray}82.71 & \cellcolor{lightgray}90.04 & \cellcolor{lightgray}98.48 & \cellcolor{lightgray}98.61 & \cellcolor{lightgray}57.30 & \cellcolor{lightgray}77.14 & \cellcolor{lightgray}80.18 &  & \cellcolor{lightgray}80.41 & \cellcolor{lightgray}68.93 & \cellcolor{lightgray}97.45 & \cellcolor{lightgray}96.69 & \cellcolor{lightgray}80.28 & \cellcolor{lightgray}92.01 & \cellcolor{lightgray}85.84 & \cellcolor{lightgray}87.46 & \cellcolor{lightgray}58.80 & \cellcolor{lightgray}75.55 & \cellcolor{lightgray}78.36 &  \\ \hline
\multirow{10}{*}{\rotatebox{90}{multimodal}} & \multirow{5}{*}{\begin{tabular}[c]{@{}c@{}}FatFormer \\ \cite{liu2024forgery}\end{tabular}} & $\mathcal{A}$ & 95.76 & 94.93 & 99.37 & 99.65 & 99.35 & 99.08 & 99.38 & 99.50 & 67.19 & 80.16 & 81.58 & \cellcolor{lightgray}92.36 & 86.89 & 87.93 & 99.30 & 99.55 & 99.40 & 99.23 & 93.22 & 94.54 & 64.76 & 82.07 & 82.07 & \cellcolor{lightgray}89.91 \\
 &  & $\mathcal{B}$ & 93.28 & 87.40 & 99.70 & 99.99 & 99.83 & 99.87 & 98.85 & 99.92 & 72.18 & 76.35 & 86.02 & \cellcolor{lightgray}92.13 & 81.56 & 75.09 & 99.60 & 99.99 & 99.84 & 99.89 & 88.35 & 99.02 & 67.40 & 75.13 & 87.01 & \cellcolor{lightgray}88.44 \\
 &  & $\mathcal{C}$ & / & 94.25 & 99.88 & 99.99 & 98.87 & 97.36 & 99.86 & 99.90 & 67.82 & 82.72 & 87.38 & \cellcolor{lightgray}92.80 & / & 85.59 & 99.85 & 99.99 & 98.43 & 97.30 & 98.32 & 98.93 & 63.71 & 83.13 & 88.25 & \cellcolor{lightgray}\textbf{91.35} \\
 &  & $\mathcal{D}$ & 96.16 & / & 99.76 & 99.78 & 98.32 & 99.74 & 99.89 & 99.66 & 65.85 & 97.44 & 80.96 & \cellcolor{lightgray}93.76 & 86.90 & / & 99.66 & 99.70 & 97.66 & 99.72 & 98.56 & 96.30 & 65.40 & 97.59 & 83.77 & \cellcolor{lightgray}92.53 \\
 &  & \cellcolor{lightgray}ave & \cellcolor{lightgray}95.07 & \cellcolor{lightgray}92.47 & \cellcolor{lightgray}\textbf{99.68} & \cellcolor{lightgray}99.85 & \cellcolor{lightgray}99.09 & \cellcolor{lightgray}99.01 & \cellcolor{lightgray}99.50 & \cellcolor{lightgray}99.75 & \cellcolor{lightgray}68.26 & \cellcolor{lightgray}84.17 & \cellcolor{lightgray}83.99 & & \cellcolor{lightgray}85.12 & \cellcolor{lightgray}82.87 & \cellcolor{lightgray}\textbf{99.60} & \cellcolor{lightgray}99.81 & \cellcolor{lightgray}98.83 & \cellcolor{lightgray}99.04 & \cellcolor{lightgray}94.61 & \cellcolor{lightgray}97.20 & \cellcolor{lightgray}65.32 & \cellcolor{lightgray}84.48 & \cellcolor{lightgray}\textbf{85.28} & \\ \cline{2-27}
 & \multirow{5}{*}{\textbf{Ours}} & $\mathcal{A}$ & 97.56 & 96.21 & 99.15 & 99.86 & 99.79 & 99.56 & 99.68 & 99.83 & 71.72 & 83.38 & 74.13 & \cellcolor{lightgray}\textbf{92.81} & 91.76 & 90.67 & 99.08 & 99.82 & 99.80 & 99.62 & 96.28 & 98.00 & 68.74 & 84.89 & 74.42 & \cellcolor{lightgray}\textbf{91.19} \\
 &  & $\mathcal{B}$ & 91.82 & 87.86 & 99.78 & 99.99 & 99.96 & 99.98 & 98.99 & 99.93 & 73.73 & 73.15 & 89.67 & \cellcolor{lightgray}\textbf{92.26} & 77.52 & 76.09 & 99.70 & 99.99 & 99.96 & 99.98 & 89.73 & 99.21 & 69.17 & 71.83 & 90.71 & \cellcolor{lightgray}\textbf{88.54} \\
 &  & $\mathcal{C}$ & / & 93.79 & 99.75 & 99.99 & 99.27 & 98.40 & 99.76 & 99.83 & 66.15 & 84.71 & 87.94 & \cellcolor{lightgray}\textbf{92.96} & / & 84.49 & 99.69 & 99.99 & 98.98 & 98.39 & 97.33 & 98.18 & 62.02 & 84.94 & 85.95 & \cellcolor{lightgray}91.00 \\
 &  & $\mathcal{D}$ & 97.98 & / & 99.89 & 99.91 & 97.03 & 99.94 & 99.92 & 99.97 & 66.70 & 99.18 & 85.76 & \cellcolor{lightgray}\textbf{94.63} & 92.96 & / & 99.84 & 99.88 & 95.59 & 99.94 & 99.10 & 99.74 & 66.29 & 99.09 & 84.36 & \cellcolor{lightgray}\textbf{93.68} \\
 &  & \cellcolor{lightgray}ave & \cellcolor{lightgray}\textbf{95.79} & \cellcolor{lightgray}\textbf{92.62} & \cellcolor{lightgray}99.64 & \cellcolor{lightgray}\textbf{99.94} & \cellcolor{lightgray}99.01 & \cellcolor{lightgray}\textbf{99.47} & \cellcolor{lightgray}\textbf{99.59} & \cellcolor{lightgray}\textbf{99.89} & \cellcolor{lightgray}\textbf{69.58} & \cellcolor{lightgray}\textbf{85.11} & \cellcolor{lightgray}\textbf{84.38} &  & \cellcolor{lightgray}\textbf{87.41} & \cellcolor{lightgray}83.75 & \cellcolor{lightgray}99.58 & \cellcolor{lightgray}\textbf{99.92} & \cellcolor{lightgray}98.58 & \cellcolor{lightgray}\textbf{99.48} & \cellcolor{lightgray}\textbf{95.61} & \cellcolor{lightgray}\textbf{98.78} & \cellcolor{lightgray}66.56 & \cellcolor{lightgray}\textbf{85.19} & \cellcolor{lightgray}83.86 &  \\
 \noalign{{\color{black}\hrule height 1pt}}
\end{tabular}
\vspace{-2mm}
\end{table*}

%% file: Tables/results_unifd.tex
\begin{table*}[htbp]
\centering
\tabcolsep=0.11cm
\caption{{Generalization results on UniFD dataset\cite{ojha2023towards}}}
\vspace{-2mm}
\label{tab:general_unifd_ap}
\fontsize{6.3pt}{6.7pt}\selectfont %
\begin{tabular}{lcccccccccccccccc}
\toprule
   &
  \multicolumn{6}{c}{Generative adversarial networks} &
   &
  \multicolumn{7}{c}{Diffusion models} & &
  Total \\ \cmidrule(lr){2-7} \cmidrule(lr){9-16} \cmidrule(lr){17-17}
   &
   &
   &
   &
   &
   &
   &
   &
   &
  \multicolumn{3}{c}{LDM} &
  \multicolumn{3}{c}{GLIDE} &
   &
   \\ \cmidrule(lr){10-12} \cmidrule(lr){13-15}
  \multirow{-4}{*}{Method} &
  \multirow{-3}{*}{\begin{tabular}[c]{@{}c@{}}ProGAN\end{tabular}} &
  \multirow{-3}{*}{\begin{tabular}[c]{@{}c@{}}CycleGAN\end{tabular}} &
  \multirow{-3}{*}{\begin{tabular}[c]{@{}c@{}}BigGAN\end{tabular}} &
  \multirow{-3}{*}{\begin{tabular}[c]{@{}c@{}}StyleGAN\end{tabular}} &
  \multirow{-3}{*}{\begin{tabular}[c]{@{}c@{}}GauGAN\end{tabular}} &
  \multirow{-3}{*}{\begin{tabular}[c]{@{}c@{}}StarGAN\end{tabular}} &
  \multirow{-3}{*}{\begin{tabular}[c]{@{}c@{}}Deepfakes\end{tabular}} &
  \multirow{-3}{*}{Guided} &
  200s &
  \begin{tabular}[c]{@{}c@{}}200s w/CFG\end{tabular} &
  100s &
  \begin{tabular}[c]{@{}c@{}}100-27\end{tabular} &
  \begin{tabular}[c]{@{}c@{}}50-27\end{tabular} &
  \begin{tabular}[c]{@{}c@{}}100-10\end{tabular} &
  \multirow{-3}{*}{DALL-E} &
  \multirow{-3}{*}{mAP} \\ \midrule
    \rowcolor[HTML]{f2f0f7} 
FreqSpec~\cite{zhang2019detecting} &
  55.39 &
  \textbf{100.0} &
  75.08 &
  55.11 &
  66.08 &
  \textbf{100.0} &
  45.18 &
  57.72 &
  77.72 &
  77.25 &
  76.47 &
  68.58 &
  64.58 &
  61.92 &
  67.77 &
 69.92 \\
\rowcolor[HTML]{f7fcf0} 
CNNSpot~\cite{wang2020cnn} &
  \textbf{100.0} &
  93.47 &
  84.50 &
  99.54 &
  89.49 &
  98.15 &
  89.02 &
  73.72 &
  70.62 &
  71.00 &
  70.54 &
  80.65 &
  84.91 &
  82.07 &
  70.59 &
  83.88 \\
  \rowcolor[HTML]{f7fcf0} 
PatchForensics~\cite{patchforensics} &
  80.88 &
  72.84 &
  71.66 &
  85.75 &
  65.99 &
  69.25 &
  76.55 &
  75.03 &
  87.10 &
  86.72 &
  86.40 &
  85.37 &
  83.73 &
  78.38 &
  75.67 &
  78.75 \\
  \rowcolor[HTML]{f7fcf0} 
CoOccurrence~\cite{CoOccurrence} &
  99.74 &
  80.95 &
  50.61 &
  98.63 &
  53.11 &
  67.99 &
  59.14 &
  70.20 &
  91.21 &
  89.02 &
  92.39 &
  89.32 &
  88.35 &
  82.79 &
  80.96 &
  79.63 \\ 
  \rowcolor[HTML]{f7fcf0} 
DIRE~\cite{DIRE} &
\textbf{100.0} &
76.73 &
72.80 &
97.06 &
68.44 &
\textbf{100.0} &
98.55 &
{94.29} &
95.17 &
{95.43} &
95.77 &
{96.18} &
{97.30} &
{97.53} &
68.73 &
90.27 \\
   \rowcolor[HTML]{f7fcf0} 
UniFD~\cite{ojha2023towards} &
  \textbf{100.0} &
  99.46 &
  99.59 &
  97.24 &
  99.98 &
  99.60 &
  82.45 &
  87.77 &
  99.14 &
  92.15 &
  99.17 &
  94.74 &
  95.34 &
  94.57 &
  97.15 &
  95.89 \\ 
\rowcolor[HTML]{f7fcf0} 
MoE (ViT-L14)~\cite{moe_lora} &
\textbf{100.0} &
99.97 &
{99.93} &
99.81 &
\textbf{100.0} &
99.81 &
91.45 &
94.40 &
99.87 &
97.85 &
99.93 &
99.12 &
99.48 &
\textbf{99.23} &
99.22 &
98.67 \\ 
\rowcolor[HTML]{fff7ec} 
FatFormer ~\cite{liu2024forgery} &
\textbf{100.0} &
\textbf{100.0} &
99.90 &
97.40 &
\textbf{100.0} &
\textbf{100.0} &
97.30 &
92.00 &
99.80 &
99.10 &
{99.90} &
{99.10} &
{99.40} &
{99.20} &
\textbf{99.80} &
{98.86} \\ 
\rowcolor[HTML]{fff7ec} 
FreqNet ~\cite{tan2024frequency} &
99.92 &
99.63 &
96.05 &
99.89 &
99.71 &
98.63 &
\textbf{99.92} &
96.27 &
96.06 &
\textbf{100.0} &
62.34 &
\textbf{99.80} &
\textbf{99.78} &
96.39 &
77.78 &
94.81 \\ 
\rowcolor[HTML]{fff7ec} 
\textbf{RGB branch (Ours)} &
\textbf{100.0} &
{99.98} &
{99.94} &
{99.97} &
{99.99} &
{99.96} &
97.89 &
95.40 &
99.81 &
98.94 &
99.86 &
98.04 &
98.33 &
97.75 &
99.34 &
99.01 \\
\rowcolor[HTML]{fff7ec} 
\textbf{Freq branch (Ours)} &
99.60 &
98.77 &
97.10 &
98.87 &
96.61 &
99.94 &
98.67 &
92.51 &
99.21 &
98.22 &
99.24 &
97.04 &
97.61 &
96.67 &
94.62 &
97.64 \\
   \rowcolor[HTML]{fff7ec} 
\textbf{multimodal (Ours)} &
\textbf{100.0} &
{99.99} &
\textbf{99.99} &
\textbf{99.99} &
{99.99} &
\textbf{100.0} &
{98.58} &
\textbf{97.00} &
\textbf{99.91} &
99.51 &
\textbf{99.94} &
98.33 &
98.81 &
98.11 &
{99.60} &
\textbf{99.31} \\
   \bottomrule
\end{tabular}
\vspace{-2mm}
\end{table*}

%% file: Sections/experiments.tex
\vspace{-2mm}
\section{Experimental Results}
We provide the results to compare the developed method to existing fake image detectors.

\vspace{-2mm}
\subsection{Cross-Domain Evaluation on Group 1 $\sim$ 11}
In \cref{tab:performance_ap_auroc}, the cross-domain testing performance for different detectors trained on groups \(\mathcal{A}\), \(\mathcal{B}\), \(\mathcal{C}\), and \(\mathcal{D}\) and tested on groups 1 to 11 is provided. The table also includes the average AP and average AUROC for each of the four training groups on the 11 testing groups, as well as the average AP and average AUROC for each of the 11 testing groups on the four training groups. From \cref{tab:performance_ap_auroc}, it can be concluded that the method performs well when the spatial and spectral branches operate independently. Additionally, the design of the multimodal backbone demonstrates superior performance in detecting fake images generated by new 3D realistic methods across testing groups 1 to 11, compared to other popular fake detectors in the spatial domain, spectral domain, and multimodal domain.

\begin{figure}[b]
  \centering
  \begin{subfigure}{0.49\linewidth}
  \centering
  \includegraphics[width=1.6in]{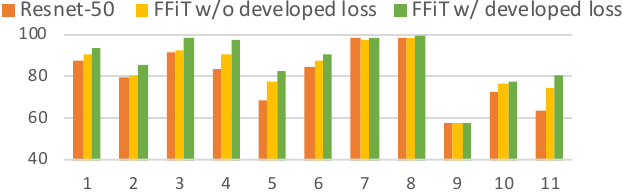}
    \caption{}
    \label{fig:ablation_ap}
  \end{subfigure}
    \begin{subfigure}{0.49\linewidth}
  \centering
\includegraphics[width=1.6in]{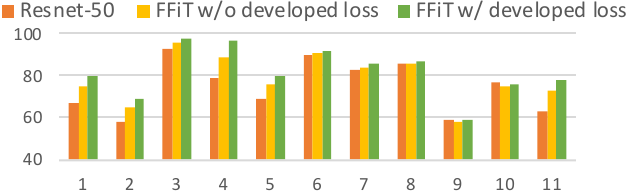}
    \caption{}
    \label{fig:ablation_auroc}
  \end{subfigure}
  \vspace{-2mm}
\caption{Ablation study of the developed loss for pre-training frequency branch, evaluated by (a) AP metric, and (b) AUROC.}
  \label{fig:ablation}
  \vspace{-2mm}
\end{figure}

\input{Tables/theory}

\subsection{Evaluation on Traditional Fake Image Dataset}
We compare the performance of the proposed method with previous methods and provide the comparative results in \cref{tab:general_unifd_ap}, quantified by the average precision metric. The results demonstrate the superior performance of our method. We follow the same training protocol as in \cite{liu2024forgery}, which involves training only on ProGAN-generated fake images, except for the method from \cite{zhang2019detecting}, which is pre-trained using CycleGAN images following \cite{moe_lora}. Notably, the methods from \cite{wang2020cnn, patchforensics, CoOccurrence, DIRE, ojha2023towards, moe_lora} are trained on 20 classes of ProGAN-generated images, while FatFormer \cite{liu2024forgery} and our method are trained on 4 classes. These differences are highlighted in \cref{tab:general_unifd_ap} with different background colors.

Our method exhibits superior performance when trained only on ProGAN images and tested on unseen GAN/DM-generated images, demonstrating its advantage in generalization to traditional generative models.

\subsection{Ablation Study on the FFiT Training Strategy}
The results of the ablation study for different training strategies of FFiT are provided in \cref{fig:ablation}, quantified by AP and AUROC, respectively. Generally, the FFiT branch pre-trained with the developed loss function outperforms the same architecture without the developed loss function. Additionally, the adopted architecture demonstrates superior performance compared to the previous architecture used in \cite{zhang2019detecting} for extracting information from the magnitude spectrum of forgery images. We further analyze the contribution of the information extracted from the spectral branch to the overall performance in the Appendix B. 

\vspace{-2mm}
\subsection{Quantitative Analysis for Generalization}
\label{sec:results_theory}
From \cref{tab:performance_ap_auroc}, it is found that the cross-domain results of the detectors trained on group \(\mathcal{A}\) always perform worse than those trained on groups \(\mathcal{B}\), \(\mathcal{C}\), and \(\mathcal{D}\), which supports the claim in \cref{sec:theory_analysis}. 

\cref{fig:distribute} presents three scenarios of the distribution projected onto a 2D space using t-SNE. The features in \cref{fig:distribute1,fig:distribute2,fig:distribute3} are extracted by three representative fake detectors: \cite{wang2020cnn}, \cite{ojha2023towards}, and the proposed method, with their representation capabilities increasing in that order. Observations reveal that the intra-cluster distances (between fake-fake clusters) remain relatively constant, while the inter-cluster distances (between real-fake clusters) increase progressively. In \cref{tab:theory}, we present the \(\rho\) values for different methods, and the values in the table support our assumption. To compute the $\rho$ in \cref{tab:theory}, we sample 2000 unseen real images from ImageNet database, and randomly sample 2000 images from training group $\mathcal{A}$, $\mathcal{B}$, $\mathcal{C}$, $\mathcal{D}$ respectively to represent the clusters of these four types of fake images during training. To simulate the unseen fake subspace, we randomly sample 200 images from testing group $1\sim 11$.

%% file: Tables/theory.tex
\begin{table}[ht]
\centering
\tabcolsep=0.066cm
\caption{The comparative $\rho$ for different methods}
\vspace{-2mm}
\label{tab:theory}
\fontsize{7.7pt}{8.9pt}\selectfont %
\begin{tabular}{ccccc}
\noalign{{\color{black}\hrule height 1pt}}
Method & CNNSpot \cite{wang2020cnn} & UniFD \cite{ojha2023towards} & FatFormer \cite{liu2024forgery} & multimodal (\textbf{ours}) \\ \hline
$\rho$ & 0.121 & 0.095 & 0.059 & 0.043 \\ 
\noalign{{\color{black}\hrule height 1pt}}
\end{tabular}
\vspace{-4mm}
\end{table}

%% file: Sections/conclusion.tex
\vspace{-2mm}
\section{Conclusions and Future Work}
This study introduces a multimodal architecture that comprehensively leverages information from both spatial and spectral domains for the detection of sophisticated fake images. To pre-train the frequency branch, we address the limitations of traditional Masked Autoencoders in handling the centrosymmetric property of spectral magnitudes by proposing a novel loss function that facilitates unsupervised learning of deep representations. In the future, the Transformer architecture trained using the developed method can be extended to other tasks that require extracting information from the spectral domain, which underscores the potential of our approach to contribute to a broader range of applications beyond synthetic content detection. The multimodal architecture, which integrates spectral and spatial features, demonstrates superior cross-domain performance even when fine-tuned on a limited number of training samples. These results confirm the approach's efficacy in accurately detecting fake images generated by advanced methods for realistic 3D scenes. Moreover, this work develops a large-scale database encompassing diverse neural-rendered images, complementing existing datasets that primarily focus on GAN and DM-generated content. Additionally, we provide a quantitative framework that can quantify the relationship between cross-domain performance and the representational capabilities of deep models, thereby enhancing the understanding of detector generalization.

This study paves the way for further advancements in the accurate detection of neural-rendered imagery while underlining the importance of robust, generalized detection for use in real-world applications.

%% file: Sections/appendix_contents.tex
\section*{\Large\textbf{\MakeUppercase{Supplementary Material}}}

\section{Squared Euclidean Distance between Two Normally Distributed Vectors}
In the main text section detailing the developed loss function for the frequency branch, we assume that the patch \(\mathbf{X}\) representing the predicted magnitude of the frequency follows a normal distribution \(\mathbf{X} \sim N(\boldsymbol{\mu}_1, \Sigma_1)\), while the patch \(\mathbf{Y}\) representing the ground truth magnitude of the frequency follows \(\mathbf{Y} \sim N(\boldsymbol{\mu}_2, \Sigma_2)\). Here, \(\boldsymbol{\mu}_1\) and \(\boldsymbol{\mu}_2\) denote the mean vectors, and \(\Sigma_1\) and \(\Sigma_2\) represent the corresponding covariance matrices.

When computing the squared Euclidean distance between these vectors, we are essentially calculating $\mathcal{L}_B = $ $(\mathbf{X} - \mathbf{Y})^\top (\mathbf{X} - \mathbf{Y})$. Letting $\mathbf{Z} = \mathbf{X} - \mathbf{Y}$, then $\mathbf{Z}$ is also a multivariate normal random vector with mean $\boldsymbol{\mu}_Z = \boldsymbol{\mu}_1 - \boldsymbol{\mu}_2$ and covariance matrix $\Sigma_Z = \Sigma_1 + \Sigma_2$ (assuming independence).

The distribution of $\mathbf{Z}^\top \mathbf{Z}$ follows a generalized chi-squared distribution. Specifically, if $\boldsymbol{\mu}_1 = \boldsymbol{\mu}_2$ and $\Sigma_1 = \Sigma_2 = I$, where $I$ is the identity matrix, then $\mathbf{Z}^\top \mathbf{Z}$ would follow a standard chi-squared distribution with $d$ degrees of freedom ($d$ being the dimension of $\mathbf{X}$ and $\mathbf{Y}$). However, when $\boldsymbol{\mu}_1 \neq \boldsymbol{\mu}_2$ or $\Sigma_1 \neq \Sigma_2$, the distribution of $\mathbf{Z}^\top \mathbf{Z}$ is a noncentral chi-squared distribution.

\noindent\textbf{Noncentral Chi-Squared Distribution}:
For $\mathbf{Z}^\top \mathbf{Z}$, the degrees of freedom $k$ equals the dimension of $\mathbf{Z}$, and the noncentrality parameter $\lambda$ is given by:
\[
\lambda = \boldsymbol{\mu}_Z^\top \Sigma_Z^{-1} \boldsymbol{\mu}_Z
\]

Thus, the distribution can be written as:
\[
\mathbf{Z}^\top \mathbf{Z} \sim \chi^2(k, \lambda)
\]

The probability density function (PDF) of a noncentral chi-squared distribution with $k$ degrees of freedom and noncentrality parameter $\lambda$ is given by:
\[
f(x; k, \lambda) = \frac{1}{2} e^{-(x+\lambda)/2} \left(\frac{x}{\lambda}\right)^{(k/4-1/2)} I_{k/2-1}(\sqrt{\lambda x}) 
\]
where $I_\nu(z)$ is the modified Bessel function of the first kind of order $\nu$.

Let $\mathcal{L}_B$ be a variable that follows the noncentral chi-squared distribution defined above, therefore, $\mathbb{E}[(1-\mathcal{L}_B)^\gamma \log \mathcal{L}_B]$ can be computed as:

\[
\int_0^\infty (1-x)^\gamma \log(x) f(x;k,\lambda) \, \mathrm{d}x
\]

\section{Contribution of Spectral Branch to the Whole Multimodal Arhictecture}
In \cref{tab:significance_analysis}, the proportional growth in performance after introducing the spectral branch into the spatial branch is computed. For each testing group from 1 to 11, 1000 features of the images from that group, extracted using the spatial branch fine-tuned on groups \(\mathcal{A}\), \(\mathcal{B}\), \(\mathcal{C}\), and \(\mathcal{D}\), are randomly sampled, resulting in a total of 4000 features. This process is repeated using the spectral branch to acquire another 4000 features. The Maximum Mean Discrepancy (MMD) score is then computed to estimate the correlation between the features extracted from the two domains.

For each testing group, the introduction of the spectral branch brings varying improvements in performance. To quantify the significance of these improvements, we compute the Pearson correlation coefficient and the corresponding $p$-value between the MMD scores and the increase in performance. Our analysis reveals a strong relationship between these two variables, indicating that the spectral branch provides independent and complementary information to the spatial branch. The greater independence of the features extracted by the spectral branch contributes to a significant improvement in multimodal performance.

\input{Tables/significance}

\section{Details of the Dataset and Protocol} \label{sec:appendix_dataset}
\subsection{Comparison of Our Database with Others}
We list the comparison of our database with the previous popular fake detection database in \cref{tab:database_comparison}.

Notably, due to the generation of 3D scenes and then project to 2D cost a lot and needed to be built from scratch, our generation cost evaluated by GPU hours is four times longer than GenImage \cite{zhu2024genimage}.

\input{Tables/database_compare}

\input{Tables/in2n_igs2gs}

\subsection{Details of Our Collected Dataset}
\input{Tables/database_collection}
We describe the details of our collected dataset in \cref{tab:details_train_data}. From \textbf{A} to \textbf{O}, we list the sources of the 2D image groups used to generate the 3D scenes. 

\subsection{Split Protocol of the Training Dataset}
\noindent\textbf{$\mathcal{A}$:} For the real images, we randomly sample 20,000 images from each folder of \{afhq, celebahq, lsun\} of ArtiFact \cite{artifactDB} database, respectively, and collect all the 4,318 images from landscape folder and all the 1,336 images from metfaces folder. Therefore we acquire a total of 65,654 real images. For GAN-generated fake images, we collect 10k, 10k, 7k, 10k, 15k, and 15k images from the folders of BigGAN, Gans-former, GauGAN, ProjectedGAN, StyleGAN3, and Taming Transformer, respectively.

\noindent\textbf{$\mathcal{B}$:} The real images are the same as $\mathcal{A}$ while a total of 66,896 fake images generated by 6 DMs are selected. Exactly, we collect 10k, 896, 20k, 6k, 20k, and 10k images from the folders of Glide, DDPM, Latent Diffusion, Palette, Stable Diffusion, and VQ Diffusion, respectively. 

\noindent\textbf{$\mathcal{C}$:} For the real class, we use all the 69,377 real images in \textbf{A}$\sim$\textbf{J}. For the rendered class, we sample 12,000 images in \textbf{A}$\sim$\textbf{J} for method \textbf{\uppercase\expandafter{\romannumeral1}},   
\textbf{\uppercase\expandafter{\romannumeral2}}, 
\textbf{\uppercase\expandafter{\romannumeral3}}, \textbf{\uppercase\expandafter{\romannumeral4}}, \textbf{\uppercase\expandafter{\romannumeral5}}, respectively. Therefore, we acquire a total of 60,000 rendered images.

\noindent\textbf{$\mathcal{D}$:} For the real class, we use all the 69,377 real images in \textbf{A}$\sim$\textbf{J}. For the rendered class, we use all the 40,734 splatfacto-rendered images in \textbf{A}$\sim$\textbf{J} and all the 10,785 C3dgs-rendered images in \textbf{G}. Therefore, we acquire a total of 51,519 rendered images.

\subsection{Details of Training the Architecture}
We pre-train the FFiT using the ImageNet dataset by resizing the images to $256\times 256$ and then cropping them to $224\times 224$. During the pre-training, we set the focusing parameter $\gamma$ in the developed loss function to 2. The learning rate is set to $1\times 10^{-4}$ with a batch size of 256 on a single H100 GPU. We employ early stopping, terminating the training when the reconstruction loss stagnates and does not improve for 5 consecutive epochs.

The spatial branch is initialized with ViT weights pre-trained on the ImageNet dataset, except for the Ada-LoRA modules, which are randomly initialized. The spectral branch is initialized with ViT weights pre-trained by FFiT, which were acquired before, with its Ada-LoRA modules also randomly initialized. Both branches are fine-tuned using a learning rate of \(1 \times 10^{-4}\) and a batch size of 256 on a single H100 GPU for 20 epochs, employing the BCEWithLogits loss. The fine-tuning follows the training protocols of groups \(\mathcal{A}\), \(\mathcal{B}\), \(\mathcal{C}\), and \(\mathcal{D}\). After acquiring the fine-tuned spatial and spectral branches, we fix their parameters and fine-tune only the last GMU layer with the FC layer to obtain the optimal parameters for fake image classification. During each training stage, including the fine-tuning of the spatial branch, spectral branch, and GMU module, we use a class-balanced random sampler, following the approach described in \cite{epstein2023online}, to balance the distribution of generated and real images over an epoch.

\subsection{Split protocol of Testing Dataset}
For testing groups 1 and 2, we select the scenes and the corresponding 2D images that never occur in the training dataset. The details are provided as follows:

\noindent{Group 1 (\textbf{\uppercase\expandafter{\romannumeral1}} $\sim$ \textbf{\uppercase\expandafter{\romannumeral5}}):} For the real class, we sample all the real images from \textbf{K}, \textbf{L}, \textbf{M}, \textbf{N}, \textbf{O}. For the fake class, we sample all the images from \textbf{K}, \textbf{L}, \textbf{M}, \textbf{N}, \textbf{O} rendered by the method \textbf{\uppercase\expandafter{\romannumeral1}}, \textbf{\uppercase\expandafter{\romannumeral2}}, \textbf{\uppercase\expandafter{\romannumeral3}}, \textbf{\uppercase\expandafter{\romannumeral4}}, \textbf{\uppercase\expandafter{\romannumeral5}}, respectively.

\noindent{Group 2 (\textbf{\uppercase\expandafter{\romannumeral5}} $\sim$ \textbf{\uppercase\expandafter{\romannumeral8}}):} For the real class, we sample all the real images from \textbf{K}, \textbf{L}, \textbf{M}, \textbf{N}, \textbf{O}. For the fake class, we sample all the images from \textbf{K}, \textbf{L}, \textbf{M}, \textbf{N}, \textbf{O} rendered by the method \textbf{\uppercase\expandafter{\romannumeral6}}, \textbf{\uppercase\expandafter{\romannumeral7}}, \textbf{\uppercase\expandafter{\romannumeral8}}, respectively.

Besides evaluating the performance of unseen fake images generated by NeRF or 3DGS, further consideration is given to scenarios where traditional generative methods, such as GANs and DMs, are combined with neural rendering techniques in groups 3, 4, 5, and 6. Additionally, the use of editable neural rendering methods is explored. In groups 7 and 8, two representative methods capable of editing 3D scenes within their 3D representations are selected. A series of prompts for 3D editing are used, and these edited 3D scenes are then projected into 2D to acquire the fake images. Another important application of neural rendering technologies, digital human (avatar) generation, is also considered. In groups 9 and 10, these technologies are used to generate images of avatars, including both heads and full bodies. In group 11, frames sampled from Sora-generated videos, which exhibit realistic 3D representations within the video, are collected.

\noindent\textbf{Pix2NeRF\cite{pix2nerf}:} For real class, We use all the 70,000 images in ffhq folder of ArtiFact \cite{artifactDB} database. For fake class, we render 96,000 ($1,000\times 96$) images, where we reconstruct 1000 identities and render 96 images from different views for each identity.

\noindent\textbf{SketchFaceNeRF\cite{sketchfacenerf}:} For real class, We use all the 70,000 images in ffhq folder of ArtiFact \cite{artifactDB} database. For fake class, we render 90,000 ($60\times 60 \times 25$) images, where we use 60 sketches to style-transfer 60 identities and render 25 images from different views for each style-transferred head.

\noindent\textbf{DreamFusion\cite{dreamfusion}:} For real class, we randomly sample 10,000 images in imagenet folder of ArtiFact \cite{artifactDB} database. For the fake class, we render 10,600 ($106\times 100$) images, where we use 106 prompts for generation and render 100 images from different views for each generated 3D scene.

\noindent\textbf{GSGEN\cite{gsgen}:} For real class, we randomly sample 10,000 images in imagenet folder of ArtiFact \cite{artifactDB} database. For the fake class, we render 9,540 ($106\times 90$) images, where we use 106 prompts for generation and render 90 images from different views for each generated 3D scene.

\noindent\textbf{Instruct-N2N\cite{instruct_n2n}:} For real class, we collect all the 3,174 real images which are used to successfully train the nerfacto (\textbf{\uppercase\expandafter{\romannumeral3}}) of dataset \textbf{K}, \textbf{L}, \textbf{M}, \textbf{N}, \textbf{O}. For the fake class, we generate 40,559 edited images from the nerfacto-generated 3D scenes. The details of this Instruct-N2N dataset can be found in \cref{tab:in2n_igs2gs}.

\noindent\textbf{Instruct-GS2GS\cite{instruct_gs2gs}:} For real class, we collect all the 3,174 real images which are used to successfully train the splatfacto (\textbf{\uppercase\expandafter{\romannumeral7}}) of dataset \textbf{K}, \textbf{L}, \textbf{M}, \textbf{N}, \textbf{O}. For the fake class, we generate 40,559 edited images from the splatfacto-generated 3D scenes. The details of this Instruct-GS2GS dataset can be found in \cref{tab:in2n_igs2gs}.

\noindent\textbf{GeneFace++\cite{genefacepp}:} We have 14 videos of different identities for training the 3D representation of speaking head. For fake speech video generation, we enable each identity to speak the contents from the other 13 identities by inputting the extracted audio, and therefore generate 182 ($14\times 13$) fake videos. For the real class, We evenly sample 650 frames from each real video and generate 9,100 images ($650\times 14$). For the fake class, we evenly sample 50 frames from each fake video and generate 9,100 images ($50 \times 182$).

\noindent\textbf{SplattingAvatar\cite{splattingavatar}:} For the real class, we use all the 33,728 images of 14 identities (10 identities are head and 4 identities are full body) provided by \cite{splattingavatar}. For the fake class, we generate 33,715 rendered images for 14 identities.

\noindent\textbf{SORA\cite{sora} frames:} For real class, we randomly sample 60,000 images in coco folder of ArtiFact \cite{artifactDB} database. For the fake class, we collected 94 publicly released videos generated by SORA and randomly cropped a total of 60,531 images in the size of $512\times 512$ pixels from the frames of these SORA-generated videos. 

\noindent\textbf{106 Prompts for Stable-DreamFusion and GSGEN:} Acropolis of Athens, Desert cactus, Jamaican jerk chicken, Red panda, African elephant, Dolphin, Japanese ramen, Redwood forest, African lion, Dutch pancakes, Japanese sushi, Rhinoceros, Alpine meadow, Egyptian koshari, King cobra, Rose garden, Amazon jungle, Eiffel Tower, Koala bear, Russian borscht, American burger, Emperor penguin, Korean barbecue, Sagrada Familia, Angkor Wat, Ethiopian injera, Korean bibimbap, Saint Basil Cathedral, Arctic wolf, French bakery, Lavender fields, Siberian tiger, Argentine steak, French crepes, Lebanese falafel, Snow leopard, Australian steak, German sausages, Machu Picchu, Spanish tapas, Bald eagle, Giant panda, Malaysian satay, Statue of Liberty, Bamboo forest, Giraffe, Maple tree, Sunflower field, Belgian waffles, Golden Gate Bridge, Mexican churros, Swedish meatballs, Bengal tiger, Gray wolf, Mexican tacos, Swiss chocolate, Blue whale, Great Barrier Reef, Moroccan couscous, Sydney Opera House, Bonsai tree, Great Wall of China, Neuschwanstein Castle, Taj Mahal, Brandenburg Gate, Great white shark, Notre Dame Cathedral, Thai curry, Brazilian barbecue, Greek salad, Oak tree, Thai mango sticky rice, British fish and chips, Grizzly bear, Orca whale, Tower Bridge, Burj Khalifa, Hagia Sophia, Orchid garden, Tropical rainforest, Canadian poutine, Hawaiian poke bowl, Palm tree, Tulip garden, Cheetah, Hippopotamus, Peruvian ceviche, Turkish kebab, Cherry blossom tree, Indian curry, Petra Jordan, Venus flytrap, Chimpanzee, Indian samosas, Pine forest, Water lily pond, Chinese dumplings, Irish stew, Polar bear, Westminster Abbey, Colosseum, Italian gelato, Red fox, Coral reef, Italian pasta, Red kangaroo.

\section{Train in the past and test in the future}
\label{sec:train_past_test_future}

\input{Tables/protocol_NN_GG}

We especially define two self-evaluation protocols (PT-\textcolor{red}{$\text{NN}$} and PT-\textcolor{red}{$\text{GG}$}) to observe the performance of detectors which is trained on the past and tested on the future: \romannumeral 1 ) train on NeRF-rendered images and test on unseen NeRF-rendered images, \romannumeral 2 ) train on 3DGS-rendered images and test on unseen 3DGS-rendered images. For PT-\textcolor{red}{$\text{NN}$} and PT-\textcolor{red}{$\text{GG}$}, the list of evaluations is provided in the \cref{tab:protocol_NN_GG}. 

\noindent\textbf{{PT-\textcolor{red}{$\text{NN}$}}}:
To train the detectors, we select the images from the methods of real, \textbf{\uppercase\expandafter{\romannumeral1}}, \textbf{\uppercase\expandafter{\romannumeral2}}, \textbf{\uppercase\expandafter{\romannumeral3}}, \textbf{\uppercase\expandafter{\romannumeral4}}, and \textbf{\uppercase\expandafter{\romannumeral5}}, excluding those from folders designated for evaluation, which are summarized in the list of evaluation. To evaluate the performance of the detectors, we utilize all images from the methods of real, \textbf{\uppercase\expandafter{\romannumeral1}}, \textbf{\uppercase\expandafter{\romannumeral2}}, \textbf{\uppercase\expandafter{\romannumeral3}}, \textbf{\uppercase\expandafter{\romannumeral4}}, and \textbf{\uppercase\expandafter{\romannumeral5}} located within the folders in the following list that is specified for evaluation: nerfstudio/\{bww-entrance, campanile, desolation, Egypt, kitchen, library, person, redwoods2, storefront, stump, vegetation\}, record3d/bear, head/face, eyefultower/\{office1b, office-view2, riverview\}, mip360/\{bicycle, bonsai, counter, flowers, kitchen, room, stump, treehill\}.

\noindent\textbf{{PT-\textcolor{red}{$\text{GG}$}}}
To train the detectors, we select the 1,786, 1,721, 1,234, 1,721 images from the \textbf{L}, \textbf{M}, \textbf{N}, \textbf{O} datasets for the methods of real, \textbf{\uppercase\expandafter{\romannumeral6}}, \textbf{\uppercase\expandafter{\romannumeral7}}, and \textbf{\uppercase\expandafter{\romannumeral8}}. To evaluate the performance of the detectors, we use the images from the  \textbf{K} dataset for the methods of real, \textbf{\uppercase\expandafter{\romannumeral6}}, \textbf{\uppercase\expandafter{\romannumeral7}}, and \textbf{\uppercase\expandafter{\romannumeral8}}.

We provide the evaluation performance on PT-\textcolor{red}{$\text{NN}$} and PT-\textcolor{red}{$\text{GG}$} of different detectors in \cref{tab:pt_nn_performance} and \cref{tab:pt_gg_performance}, respectively. We sort the neural rendering methods according to the release date, for example, "$\leq$ nerfacto" means we use the fake images generated by i-ngp, tensorf and nerfacto for training. The results in \cite{epstein2023online} reveal that the testing performance on the recently released generation methods of the detector can be benefit from more exposure to fake images generated by newly developed methods during training. However, observed from \cref{tab:pt_nn_performance} and \cref{tab:pt_gg_performance}, such trend does not show in NeRF and 3DGS-generated fake image detection.

\input{Tables/results_pt_nn}

\input{Tables/results_pt_gg}

\section{Explaination on Evaluation Protocols}
We select several representative methods for comparison. We don't compare with NPR \cite{NPR} since literature \cite{cozzolino2024raising} gives a fair comparison of UniFD \cite{ojha2023towards} with NPR \cite{NPR} and gives the conclusion that UniFD \cite{ojha2023towards} is better than NPR \cite{NPR}. We don't compare with \cite{cozzolino2024raising} since this method utilizes the text extractor to extract the text description from real images, and then input the text to a diffusion model-based generator to synthesize fake images. Then SVM classifier is trained based on real and such synthesized images. However we cannot use the same way to synthesize training samples by Nerf/3DGS methods for group $\mathcal{C}$ and $\mathcal{D}$. We don't compare with \cite{cazenavette2024fakeinversion,zhou2024stealthdiffusion,ricker2024aeroblade} since they utilize the inherent features of DM and they are DM-only methods.

\subsection{Evaluation Metrics}
The evaluation performance, quantified by average precision (AP) and AUROC, of the benchmark detectors and the proposed method is provided in \cref{tab:performance_ap_auroc}. The performance of the proposed method is compared with re-implemented detectors \cite{wang2020cnn, ojha2023towards, moe_lora} for the spatial branch only, and with the re-implemented detector \cite{zhang2019detecting} for the spectral branch only. The multimodal backbone is compared with FatFormer \cite{liu2024forgery} by re-implementing the spatial-spectral multimodal branch of \cite{liu2024forgery} while removing the language-based branch for fair comparison. This adjustment not only streamlines the comparison but also mitigates discrepancies in the text-guided interaction \cite{liu2024forgery} module during evaluation, given the partial and low-level scenes prevalent in the NeRF/3DGS-generated images of the testing dataset, unlike the uniform rich contexts of GAN/DM outputs used for testing in \cite{liu2024forgery}. When comparing the testing results on GAN/DM-generated images with detectors trained on ProGAN images only, the proposed methods are compared with FatFormer \cite{liu2024forgery} while keeping their language branch.

For a rigorous cross-domain evaluation, the performance metrics of detectors that are both trained and tested on images generated by the same method, such as those produced by NeRF or 3DGS, are not included in the primary results. These cases are considered within-domain evaluations and do not align with the cross-domain assessment objectives. Instead, the findings from these within-domain tests are detailed in Sec. \ref{sec:train_past_test_future}, adhering to a separate evaluation protocol.

%% file: Tables/significance.tex
\begin{table*}[ht]
\centering
\tabcolsep=0.17cm
\fontsize{8pt}{8.4pt}\selectfont %
\caption{Significance of the Spectral Branch}
\label{tab:significance_analysis}
\begin{tabular}{ccccccccccccc|c}
\noalign{{\color{black}\hrule height 1pt}}
\multicolumn{2}{c}{} & 1 & 2 & 3 & 4 & 5 & 6 & 7 & 8 & 9 & 10 & 11 & \multirow{2}{*}{\begin{tabular}[c]{@{}c@{}}Pearson\\ Correlation:\end{tabular}} \\ \cline{1-13}
\multirow{2}{*}{\begin{tabular}[c]{@{}c@{}}Average Precision\\ (in \%)\end{tabular}} & \ding{172} spatial only & 93.50 & 91.68 & 87.13 & 96.02 & 99.44 & 98.76 & 98.98 & 99.41 & 68.21 & 59.92 & 57.63 &  \\
 & \ding{173} multimodal & 95.79 & 92.62 & 99.64 & 99.94 & 99.01 & 99.47 & 99.59 & 99.89 & 69.58 & 85.11 & 84.38 & 0.8816 \\ \cline{1-13}
\multicolumn{2}{c}{(\ding{173} $-$ \ding{172})/\ding{172} (in \%)} & 2.45 & 1.03 & 14.36 & 4.08 & -0.43 & 0.72 & 0.62 & 0.48 & 2.01 & 42.04 & 46.42 & $p$-value: \\
\multicolumn{2}{c}{MMD(spatial,spectral)} & 4.74 & 3.05 & 6.38 & 5.66 & 2.27 & 3.51 & 3.62 & 2.97 & 4.13 & 7.54 & 7.89 & 3.32$\times 10^{-4}$ \\ 
\noalign{{\color{black}\hrule height 1pt}}
\end{tabular}
\end{table*}

%% file: Tables/database_compare.tex
\begin{table*}[ht]
    \centering
    \tabcolsep=0.4cm
\begin{threeparttable}
\fontsize{7pt}{7.5pt}\selectfont %
    \caption{Comparison of different fake databases.}
\begin{tabular}{cccccc}
\noalign{{\color{black}\hrule height 1pt}}
\multirow{2}{*}{Method} & \multirow{2}{*}{Task Type} & \multicolumn{2}{c}{Method} & \multirow{2}{*}{\textit{\textbf{real:fake}}} & \multirow{2}{*}{releasing year} \\ \cline{3-4}
 &  & generative model & neural rendering &  &  \\ \hline
DFFD \cite{dffd} & deepfake face & GAN & \ding{55} & 58,703: 240,336 & 2020 \\
ForgeryNet \cite{he2021forgerynet} & deepfake face & GAN & \ding{55} & 1,438,201: 1,457,861 & 2021 \\
DeepArt \cite{wang2023benchmarking} & deepfake art & DM & \ding{55} & 64,479: 73,411 & 2023 \\ \hline
CNNSpot \cite{wang2020cnn} & general & GAN \tnote{1} & \ding{55} & 362,000: 362,000 & 2020 \\
CIFAKE \cite{bird2024cifake} & general & DM \tnote{2} & \ding{55} & 60,000: 60,000 & 2023 \\
UniFD \cite{ojha2023towards} & general & DM \tnote{3} & \ding{55} & 1000: 8000 & 2023 \\
GenImage \cite{zhu2024genimage} & general & GAN, DM \tnote{4} & \ding{55} & 1,331,167: 1,350,000 & 2023 \\
\textit{\textbf{ours}} & general & Sora & NeRF, 3DGS & 259,105: 687,108 & 2024 \\ 
\noalign{{\color{black}\hrule height 1pt}}
\end{tabular}
\begin{tablenotes}
    \item[1] ProGAN, CycleGAN, BigGAN, StyleGAN, GauGAN, StarGAN, Deepfakes, SITD, SAN, CRN, IMLE 
    \item[2] Stable Diffusion Model
    \item[3] Glide, LDM, Dalle-mini
    \item[4] GAN (BigGAN) \& DM (Midjourney SD1.4, Midjourney SD1.5, ADM, Glide, Wukong, VQDM) 
\end{tablenotes}
\label{tab:database_comparison}
\end{threeparttable}
\end{table*}

%% file: Tables/in2n_igs2gs.tex
\begin{table*}[ht]
\centering
\tabcolsep=0.26cm
\fontsize{7pt}{7.4pt}\selectfont %
\caption{The Detailed Information of Our Instruct-N2N and Instruct-GS2GS Dataset. (generated scenes/rendered images)}
\label{tab:in2n_igs2gs}
\begin{tabular}{c|cccccc|ccccc}
\noalign{{\color{black}\hrule height 1pt}}
\multirow{2}{*}{} & \multirow{2}{*}{Prompts} & \multicolumn{5}{c|}{instruct-N2N} & \multicolumn{5}{c}{instruct-GS2GS} \\ \cline{3-12} 
 &  & \textbf{K} \cite{mipnerf360} & \textbf{L} \cite{llff} & \textbf{M} \cite{instruct_n2n,nerfart} & \textbf{N} \cite{3dgs} & \textbf{O} \cite{seathru} & \textbf{K} \cite{mipnerf360} & \textbf{L} \cite{llff} & \textbf{M} \cite{instruct_n2n,nerfart} & \textbf{N} \cite{3dgs} & \textbf{O} \cite{seathru} \\ \hline
\multirow{19}{*}{\rotatebox{90}{prompts for human}} & Indian attire & \ding{55}& \ding{55}& 4/353 & \ding{55}& \ding{55}& \ding{55}& \ding{55}& 4/353 & \ding{55}& \ding{55}\\
 & Mustache & \ding{55}& \ding{55}& 4/353 & \ding{55}& \ding{55}& \ding{55}& \ding{55}& 4/353 & \ding{55}& \ding{55}\\
 & Bronze statue & \ding{55}& \ding{55}& 4/353 & \ding{55}& \ding{55}& \ding{55}& \ding{55}& 4/353 & \ding{55}& \ding{55}\\
 & Joker makeup & \ding{55}& \ding{55}& 4/353 & \ding{55}& \ding{55}& \ding{55}& \ding{55}& 4/353 & \ding{55}& \ding{55}\\
 & Gothic makeup & \ding{55}& \ding{55}& 4/353 & \ding{55}& \ding{55}& \ding{55}& \ding{55}& 4/353 & \ding{55}& \ding{55}\\
 & Anime eyes & \ding{55}& \ding{55}& 4/353 & \ding{55}& \ding{55}& \ding{55}& \ding{55}& 4/353 & \ding{55}& \ding{55}\\
 & Vintage sepia tone & \ding{55}& \ding{55}& 4/353 & \ding{55}& \ding{55}& \ding{55}& \ding{55}& 4/353 & \ding{55}& \ding{55}\\
 & Neon lights & \ding{55}& \ding{55}& 4/353 & \ding{55}& \ding{55}& \ding{55}& \ding{55}& 4/353 & \ding{55}& \ding{55}\\
 & Cyberpunk style & \ding{55}& \ding{55}& 4/353 & \ding{55}& \ding{55}& \ding{55}& \ding{55}& 4/353 & \ding{55}& \ding{55}\\
 & Renaissance painting & \ding{55}& \ding{55}& 4/353 & \ding{55}& \ding{55}& \ding{55}& \ding{55}& 4/353 & \ding{55}& \ding{55}\\
 & Pop art & \ding{55}& \ding{55}& 4/353 & \ding{55}& \ding{55}& \ding{55}& \ding{55}& 4/353 & \ding{55}& \ding{55}\\
 & Tribal face paint & \ding{55}& \ding{55}& 4/353 & \ding{55}& \ding{55}& \ding{55}& \ding{55}& 4/353 & \ding{55}& \ding{55}\\
 & Alien features & \ding{55}& \ding{55}& 4/353 & \ding{55}& \ding{55}& \ding{55}& \ding{55}& 4/353 & \ding{55}& \ding{55}\\
 & Pixel art & \ding{55}& \ding{55}& 4/353 & \ding{55}& \ding{55}& \ding{55}& \ding{55}& 4/353 & \ding{55}& \ding{55}\\
 & Watercolor effect & \ding{55}& \ding{55}& 4/353 & \ding{55}& \ding{55}& \ding{55}& \ding{55}& 4/353 & \ding{55}& \ding{55}\\
 & Sketch drawing & \ding{55}& \ding{55}& 4/353 & \ding{55}& \ding{55}& \ding{55}& \ding{55}& 4/353 & \ding{55}& \ding{55}\\
 & Surreal distortion & \ding{55}& \ding{55}& 4/353 & \ding{55}& \ding{55}& \ding{55}& \ding{55}& 4/353 & \ding{55}& \ding{55}\\
 & Film noir & \ding{55}& \ding{55}& 4/353 & \ding{55}& \ding{55}& \ding{55}& \ding{55}& 4/353 & \ding{55}& \ding{55}\\
 & Glitch art & \ding{55}& \ding{55}& 4/353 & \ding{55}& \ding{55}& \ding{55}& \ding{55}& 4/353 & \ding{55}& \ding{55}\\ \hline
\multirow{12}{*}{\rotatebox{90}{prompts for nature}} & Snowy landscape & 9/1,940 & 8/305 & \ding{55}& 2/488 & 4/88 & 9/1,940 & 8/305 & \ding{55}& 2/488 & 4/88 \\
 & summer style & 9/1,940 & 8/305 & \ding{55}& 2/488 & 4/88 & 9/1,940 & 8/305 & \ding{55}& 2/488 & 4/88 \\
 & Autumn foliage & 9/1,940 & 8/305 & \ding{55}& 2/488 & 4/88 & 9/1,940 & 8/305 & \ding{55}& 2/488 & 4/88 \\
 & spring style & 9/1,940 & 8/305 & \ding{55}& 2/488 & 4/88 & 9/1,940 & 8/305 & \ding{55}& 2/488 & 4/88 \\
 & Tropical paradise & 9/1,940 & 8/305 & \ding{55}& 2/488 & 4/88 & 9/1,940 & 8/305 & \ding{55}& 2/488 & 4/88 \\
 & Ancient style & 9/1,940 & 8/305 & \ding{55}& 2/488 & 4/88 & 9/1,940 & 8/305 & \ding{55}& 2/488 & 4/88 \\
 & High brightness & 9/1,940 & 8/305 & \ding{55}& 2/488 & 4/88 & 9/1,940 & 8/305 & \ding{55}& 2/488 & 4/88 \\
 & Halloween theme & 9/1,940 & 8/305 & \ding{55}& 2/488 & 4/88 & 9/1,940 & 8/305 & \ding{55}& 2/488 & 4/88 \\
 & Cosmic style & 9/1,940 & 8/305 & \ding{55}& 2/488 & 4/88 & 9/1,940 & 8/305 & \ding{55}& 2/488 & 4/88 \\
 & Industrial chic & 9/1,940 & 8/305 & \ding{55}& 2/488 & 4/88 & 9/1,940 & 8/305 & \ding{55}& 2/488 & 4/88 \\
 & cyberpunk & 9/1,940 & 8/305 & \ding{55}& 2/488 & 4/88 & 9/1,940 & 8/305 & \ding{55}& 2/488 & 4/88 \\
 & Baroque inspiration & 9/1,940 & 8/305 & \ding{55}& 2/488 & 4/88 & 9/1,940 & 8/305 & \ding{55}& 2/488 & 4/88 \\
 \noalign{{\color{black}\hrule height 1pt}}
\end{tabular}
\end{table*}

%% file: Tables/database_collection.tex
\begin{table*}[ht]
\centering
\tabcolsep=0.26cm
\caption{The Numbers ($x/y$) of Generated Scenes and the Corresponding 2D Images of Our Database: $x,y$ denote the number of generated 3D scenes and rendered 2D images, respectively. Only 3D scenes that are successfully reconstructed are adopted.}
\vspace{-2mm}
\label{tab:details_train_data}
\fontsize{7.6pt}{9pt}\selectfont %
\begin{threeparttable}
\begin{tabular}{lccccccccc}
\noalign{{\color{black}\hrule height 1pt}}
\multirow{2}{*}{\begin{tabular}[c]{@{}c@{}} Sources of \\ the Scenes \end{tabular}} &  & \textbf{\uppercase\expandafter{\romannumeral1}} \cite{ingp} & \textbf{\uppercase\expandafter{\romannumeral2}} \cite{tensorf} & \textbf{\uppercase\expandafter{\romannumeral3}} \cite{tancik2023nerfstudio} & \textbf{\uppercase\expandafter{\romannumeral4}} \cite{seathru} & \textbf{\uppercase\expandafter{\romannumeral5}} \cite{pynerf} & \textbf{\uppercase\expandafter{\romannumeral6}} \cite{3dgs} & \textbf{\uppercase\expandafter{\romannumeral7}} \cite{tancik2023nerfstudio} & \textbf{\uppercase\expandafter{\romannumeral8}} \cite{c3dgs} \\ \cline{3-10} 
 & real & i-ngp & tensorf & nerfacto & seaThru & pynerf & 3dgs & splatfacto & C3dgs \\
\hline
\textbf{A} blender \cite{tancik2023nerfstudio} & 8/800 & 8/800 & 8/800 & 6/600 & 8/800 & 8/800 & \ding{55} & 2/200 & \ding{55} \\
\textbf{B} D-NeRF \cite{tancik2023nerfstudio} & 8/1,100 & 7/1,000 & 8/1,100 & 5/550 & 8/1,100 & 8/1,100 & \ding{55} & \ding{55} & \ding{55} \\
\textbf{C} eyeful-T \cite{tancik2023nerfstudio} & 11/28,572 & 11/28,572 & 8/25,107 & 11/28,572 & 8/15,465 & 5/10,668 & \ding{55} & 7/9,654 & \ding{55} \\
\textbf{D} mill19 \cite{tancik2023nerfstudio} & 2/3,618 & 2/3,618 & 2/3,618 & 2/3,618 & 2/3,618 & 2/3,618 & \ding{55} & 2/3,618 & \ding{55} \\
\textbf{E} nerfosr \cite{tancik2023nerfstudio} & 9/4,703 & 9/4,426 & 7/3,634 & 9/4,426 & \ding{55} & \ding{55} & \ding{55} & 7/3,634 & \ding{55} \\
\textbf{F} nerfstudio \cite{tancik2023nerfstudio} & 17/6,321 & 17/6,321 & 17/6,321 & 17/6,321 & 17/6,321 & 11/4,605 & \ding{55} & 17/6,321 & \ding{55} \\
\textbf{G} phototour \cite{tancik2023nerfstudio} & 10/17,741 & \ding{55} & \ding{55} & 10/17,741 & \ding{55} & \ding{55} & \ding{55} & 8/10,785 & 8/10,785 \\
\textbf{H} record3d \cite{tancik2023nerfstudio} & 1/300 & 1/300 & 1/300 & 1/300 & 1/300 & 1/300 & \ding{55} & 1/300 & \ding{55} \\
\textbf{I} sdfStudio \cite{tancik2023nerfstudio} & 34/4,771 & 34/4,771 & 34/4,771 & 34/4,771 & 32/4,172 & 31/3,871 & \ding{55} & 34/4,771 & \ding{55} \\
\textbf{J} sitcoms3d \cite{tancik2023nerfstudio} & 10/1,451 & \ding{55} & 10/1,451 & \ding{55} & \ding{55} & \ding{55} & \ding{55} & 10/1,451 & \ding{55} \\ \hline
\textbf{K} mip360 \cite{mipnerf360} & 9/1,940 & 9/1,940 & 8/1,755 & 9/1,940 & 9/1,940 & 8/1,755 & 9/1,940 & 9/1,940 & 9/1,940 \\
\textbf{L} llff \cite{llff} & 8/305 & 7/250 & 2/45 & 8/305 & 4/105 & 5/146 & 8/305 & 8/305 & 8/305 \\
\textbf{M} \tnote{1} $\text{head}$ \cite{instruct_n2n,nerfart} & 4/353 & 4/353 & 1/65 & 4/353 & 2/160 & 1/65 & 3/288 & 4/353 & 3/288 \\
\textbf{N} inria \cite{3dgs} & 4/1,040 & 2/488 & 2/488 & 2/488 & 2/488 & 2/488 & 4/1,040 & 2/488 & 4/1,040 \\
\textbf{O} underwater \cite{seathru} & 4/88 & 4/88 & 1/20 & 4/88 & 4/88 & 2/41 & 4/88 & 4/88 & 4/88 \\ \hline
sum scenes/images & 139/73,103 & 115/52,927 & 109/49,475 & 122/70,073 & 97/34,557 & 84/27,457 & 28/3,661 & 115/43,908 & 36/14,446
 \\ \noalign{{\color{black}\hrule height 1pt}}
\end{tabular}
\begin{tablenotes}
    \item[1] The head/face is from \cite{instruct_n2n} and head/\{fangzhou,yanan,shuquan\} are from \cite{nerfart}.
\end{tablenotes}
\end{threeparttable}
\end{table*}

%% file: Tables/protocol_NN_GG.tex
\begin{table}[ht]
\centering
\tabcolsep=0.24cm
\caption{Dataset Split of PT-\textcolor{red}{$\text{NN}$} and PT-\textcolor{red}{$\text{GG}$} 
\label{tab:protocol_NN_GG}}
\fontsize{7.4pt}{10pt}\selectfont %
\begin{tabular}{ccccccc}
\noalign{{\color{black}\hrule height 1pt}}
\multicolumn{2}{c}{\multirow{2}{*}{method}} & \multirow{2}{*}{\begin{tabular}[c]{@{}c@{}}release\\ -date\end{tabular}} & \multicolumn{2}{c}{training} & \multicolumn{2}{c}{evaluation} \\
\multicolumn{2}{c}{} &  & scenes & images & scenes & images \\ \hline
\multicolumn{1}{c|}{\multirow{6}{*}{\rotatebox{90}{PT-\textcolor{red}{$\text{NN}$}}}} & real & / & 115 & 62,286 & 24 & 10,817 \\
\multicolumn{1}{c|}{} & \textbf{\uppercase\expandafter{\romannumeral1}} \cite{ingp} & 2022 Jan & 91 & 42,110 & 24 & 10,817 \\
\multicolumn{1}{c|}{} & \textbf{\uppercase\expandafter{\romannumeral2}} \cite{tensorf} & 2022 Mar & 85 & 38,658 & 24 & 10,817 \\
\multicolumn{1}{c|}{} & \textbf{\uppercase\expandafter{\romannumeral3}} \cite{tancik2023nerfstudio} & 2023 Feb & 98 & 59,256 & 24 & 10,817 \\
\multicolumn{1}{c|}{} & \textbf{\uppercase\expandafter{\romannumeral4}} \cite{seathru} & 2023 Apr & 73 & 23,740 & 24 & 10,817 \\
\multicolumn{1}{c|}{} & \textbf{\uppercase\expandafter{\romannumeral5}} \cite{pynerf} & 2023 Nov & 60 & 16,640 & 24 & 10,817 \\ \hline
\multicolumn{1}{c|}{\multirow{4}{*}{\rotatebox{90}{PT-\textcolor{red}{$\text{GG}$}}}} & real & / & 20 & 1,786 & 9 & 1,940 \\
\multicolumn{1}{c|}{} & \textbf{\uppercase\expandafter{\romannumeral6}} \cite{3dgs} & 2023 Aug & 19 & 1,721 & 9 & 1,940 \\
\multicolumn{1}{c|}{} & \textbf{\uppercase\expandafter{\romannumeral7}} \cite{tancik2023nerfstudio} & 2023 Sep & 18 & 1,234 & 9 & 1,940 \\
\multicolumn{1}{c|}{} & \textbf{\uppercase\expandafter{\romannumeral8}} \cite{c3dgs} & 2024 Feb & 19 & 1,721 & 9 & 1,940 \\ \noalign{{\color{black}\hrule height 1pt}}
\end{tabular}
\end{table}

%% file: Tables/results_pt_nn.tex
\begin{table}[ht]
\centering
\tabcolsep=0.09cm
\caption{Comparative Performance of PT-\textcolor{red}{$\text{NN}$} (AP/AUROC)}
\label{tab:pt_nn_performance}
\fontsize{6.2pt}{8.9pt}\selectfont %
\begin{tabular}{l|cccccc}
\noalign{{\color{black}\hrule height 1pt}}
\diagbox{train}{test} &  & i-ngp & tensorf & nerfacto & seaThru & pyNerf \\ \hline
$\leq$ i-ngp & \multirow{5}{*}{{\cite{wang2020cnn}}} & 96.29/ 95.76 & 98.98/ 98.97 & 95.52/ 94.93 & 97.18/ 96.82 & 97.60/ 97.39  \\
$\leq$ tensorf & & 94.63/ 93.55 & 98.49/ 98.32 & 93.35/ 92.27 & 95.94/ 95.01 & 97.00/ 96.46 \\
$\leq$ nerfacto & & 95.01/ 94.00 & 98.33/ 98.02 & 93.96/ 93.00 & 95.89/ 94.91 & 96.42/ 95.48 \\
$\leq$ seaThru & & 95.09/ 94.35 & 98.84/ 98.73 & 94.07/ 93.34 & 96.33/ 95.79 & 97.01/ 96.36 \\
$\leq$ pyNerf & & 95.18/ 95.07 & 98.69/ 98.64 & 94.12/ 94.06 & 96.19/ 95.85 & 96.63/ 96.30 \\ \hline
$\leq$ i-ngp & \multirow{5}{*}{{\cite{zhang2019detecting}}} & 93.64/ 91.97 & 96.77/ 96.31 & 93.14/ 91.53 & 92.21/ 90.67 & 92.77/ 91.67 \\
$\leq$ tensorf & & 94.41/ 92.66 & 97.74/ 97.20 & 94.01/ 92.15 & 93.40/ 91.75 & 94.12/ 92.88 \\
$\leq$ nerfacto & & 92.97/ 91.21 & 96.70/ 95.94 & 92.59/ 90.58 & 91.82/ 90.16 & 93.75/ 92.71 \\
$\leq$ seaThru & & 94.87/ 94.08 & 97.95/ 97.67 & 94.44/ 93.49 & 94.38/ 93.61 & 95.66/ 95.21 \\
$\leq$ pyNerf & & 94.81/ 93.77 & 97.20/ 96.50 & 94.32/ 93.20 & 94.20/ 93.21 & 95.33/ 94.59 \\ \hline
$\leq$ i-ngp & \multirow{5}{*}{{\cite{ojha2023towards}}} & 94.09/ 94.45 & 98.42/ 98.52 & 91.69/ 92.60 & 94.39/ 94.92 & 96.43/ 96.34 \\
$\leq$ tensorf & & 93.77/ 93.78 & 99.26/ 99.25 & 90.10/ 90.36 & 94.49/ 94.44 & 96.98/ 96.69 \\
$\leq$ nerfacto & & 92.46/ 92.77 & 98.80/ 98.78 & 88.52/ 89.23 & 93.17/ 93.34 & 96.39/ 96.14 \\
$\leq$ seaThru & & 92.38/ 92.83 & 98.78/ 98.77 & 88.27/ 89.11 & 93.38/ 93.64 & 96.42/ 96.20 \\
$\leq$ pyNerf & & 92.12/ 92.73 & 98.73/ 98.72 & 88.00/ 89.03 & 93.41/ 93.77 & 96.54/ 96.36 \\ \hline
$\leq$ i-ngp & \multirow{5}{*}{{\textbf{Ours}}} & 97.55/ 96.78 & 99.12/ 99.06 & 96.74/ 95.43 & 97.39/ 97.10 & 98.62/ 97.87 \\
$\leq$ tensorf &  & 96.24/ 94.91 & 99.14/ 99.20 & 95.32/ 93.17 & 96.84/ 96.59 & 97.21/ 97.93 \\
$\leq$ nerfacto &  & 95.42/ 95.18 & 98.97/ 98.86 & 94.73/ 93.61 & 96.54/ 95.32 & 97.89/ 96.25 \\
$\leq$ seaThru &  & 96.35/ 95.21 & 99.48/ 99.12 & 97.63/ 95.89 & 97.27/ 96.54 & 97.91/ 97.36 \\
$\leq$ pyNerf &  & 96.45/ 96.23 & 99.38/ 99.34 & 96.56/ 95.12 & 96.89/ 96.71 & 97.32/ 96.45 \\
\noalign{{\color{black}\hrule height 1pt}}
\end{tabular}
\end{table}

%% file: Tables/results_pt_gg.tex
\begin{table}[ht]
\centering
\tabcolsep=0.27cm
\caption{Comparative Performance of PT-\textcolor{red}{$\text{GG}$} (AP/AUROC)}
\label{tab:pt_gg_performance}
\fontsize{6.8pt}{9pt}\selectfont %
\begin{tabular}{l|cccc}
\noalign{{\color{black}\hrule height 1pt}}
\diagbox{train}{test} &  & 3dgs & splatfacto & c3dgs \\ \hline
$\leq$3dgs & \multirow{3}{*}{\cite{wang2020cnn}} & 79.77/ 76.73 & 73.47/ 70.05 & 81.21/ 78.52 \\
$\leq$splatfacto &  & 90.11/ 87.62 & 86.86/ 84.21 & 91.03/ 88.75 \\
$\leq$c3dgs &  & 95.34/ 94.00 & 95.18/ 94.22 & 96.09/ 94.97 \\ \hline
$\leq$3dgs & \multirow{3}{*}{\cite{zhang2019detecting}} & 65.83/ 67.52 & 64.55/ 66.09 & 69.78/ 70.89 \\
$\leq$splatfacto &  & 63.82/ 66.34 & 61.37/ 65.23 & 68.61/ 70.62 \\
$\leq$c3dgs &  & 63.33/ 65.10 & 59.25/ 63.29 & 68.44/ 69.53 \\ \hline
$\leq$3dgs & \multirow{3}{*}{\cite{ojha2023towards}} & 91.71/ 89.18 & 84.13/ 79.48 & 93.75/ 91.75 \\
$\leq$splatfacto &  & 91.13/ 88.21 & 87.00/ 82.95 & 93.40/ 91.21 \\
$\leq$c3dgs &  & 92.55/ 90.39 & 88.21/ 84.68 & 94.55/ 93.06 \\ \hline
$\leq$3dgs & \multirow{3}{*}{\textbf{Ours}} & 93.67/ 92.34 & 88.59/ 82.14 & 94.42/ 91.88 \\
$\leq$splatfacto &  & 92.76/ 91.45 & 89.32/ 85.67 & 94.58/ 93.12 \\
$\leq$c3dgs &  & 95.32/ 94.78 & 95.67/ 95.45 & 96.19/ 95.82 \\ 
\noalign{{\color{black}\hrule height 1pt}}
\end{tabular}
\end{table}